# Stochastic Primal Dual Coordinate Method with Non-Uniform Sampling Based on Optimality Violations

**Atsushi Shibagaki**                                           shibagaki.a.mllab.nit@gmail.com
*Department of Scientific and Engineering Simulation, Nagoya Institute of Technology*

**Ichiro Takeuchi**[†]                                           takeuchi.ichiro@nitech.ac.jp
*Department of Computer Science/Research Institute for Information Science, Nagoya Institute of Technology*
*RIKEN Center for Advanced Intelligence Project*
*Center for Materials Research by Information Integration, National Institute for Materials Science*

## Abstract

We study primal-dual type stochastic optimization algorithms with non-uniform sampling. Our main theoretical contribution in this paper is to present a convergence analysis of Stochastic Primal Dual Coordinate (SPDC) Method with arbitrary sampling. Based on this theoretical framework, we propose Optimality Violation-based Sampling SPDC (*ovsSPDC*), a non-uniform sampling method based on *Optimality Violation*. We also propose two efficient heuristic variants of ovsSPDC called ovsSDPC+ and ovsSDPC++. Through intensive numerical experiments, we demonstrate that the proposed method and its variants are faster than other state-of-the-art primal-dual type stochastic optimization methods.

## 1 Introduction

In this paper, we consider linear classification and regression problems by solving a regularized empirical risk minimization (ERM) problem with the number of instances $n$ and the number of features $d$ in the following form

$$\min_w \left\{ P(w) := \frac{1}{n} \sum_{i \in [n]} f_i(x_i^\top w) + \lambda g(w) \right\}, \tag{1}$$

where $f_i : \mathbb{R} \to \mathbb{R}$ is a loss function, $g : \mathbb{R}^d \to \mathbb{R}$ is a penalty function, and $\lambda > 0$ is regularization parameter. The training set is written as $\{(x_i, y_i)\}_{i \in [n]}$ where $x_i \in \mathbb{R}^d$, $y_i \in \mathbb{R}$ for regression, $y_i \in \{\pm 1\}$ for classification, and $[n] := \{1, \ldots, n\}$. We also denote $X \in \mathbb{R}^{n \times d}$ to be the data matrix defined as $X := [x_1, \ldots, x_n]^\top$ and its $j$-th column is denoted as $X_{:j}$. We assume that $f_i$ is convex and $1/\gamma$-smooth, i.e., the gradient is $1/\gamma$-Lipschitz continuous for all $i \in [n]$, $g$ is 1-strongly convex, and $\|x_i\|_2 \neq 0$ for all $i \in [n]$. As a working example of a loss function that has $1/\gamma$-smoothness, we consider smoothed hinge loss defined as $f_i(z) := 0$ for $y_i z > 1$ and $1 - y_i z - \frac{\gamma}{2}$ for $y_i z < 1 - \gamma$ and $\frac{1}{2\gamma}(1 - y_i z)^2$ otherwise. As an example of 1-strongly convex penalty, we consider elastic net penalty $g(w) := \|w\|_1 + \frac{1}{2}\|w\|_2^2$. We note that all the theories and algorithms discussed in this paper can be applied to any other combination of $1/\gamma$-smooth loss and 1-strongly convex penalty.

When the data is large, stochastic optimization is often the method of choice for solving an ERM problem in (1). Among several types of stochastic optimization methods, we focus in this paper on *primal-dual* type stochastic optimization methods (Shalev-Shwartz and Zhang, 2013; Csiba et al., 2015; Zhao and Zhang, 2015; Vainsencher et al., 2015; Qu et al., 2015; Zhang and Xiao, 2015; Zhu and Storkey, 2015; Wei Yu et al., 2015; Qu et al., 2016). The dual problem of the empirical risk minimization problem (1) is written as

$$\max_{\alpha} \left\{ D(\alpha) := -\frac{1}{n} \sum_{i \in [n]} f_i^*(-\alpha_i) - \lambda g^*\left(\frac{X^\top \alpha}{\lambda n}\right) \right\}, \tag{2}$$

where $f_i^*$ is the convex conjugate function of $f_i$, and $g^*$ is the convex conjugate function of $g$. For example, the convex conjugate of smoothed hinge loss is written as $f_i^*(z) = \frac{\gamma}{2} z^2 + y_i z$ for $y_i z \in [-1, 0]$ and $\infty$ otherwise. The convex conjugate of elastic net penalty is $g^*(v) = \frac{1}{2} \sum_{j \in [d]} (\max\{|v_j| - 1, 0\})^2$.

---
[†]Corresponding author

In primal-dual type optimization methods, the dual variables and the primal variables are alternatively updated. Most existing primal-dual type stochastic optimization methods uses the uniform sampling for selecting one or more dual variables to be updated in each iteration. Recently, several studies are conducted for non-uniform sampling (Csiba et al., 2015; Zhao and Zhang, 2015; Vainsencher et al., 2015; Qu et al., 2015; Zhang and Xiao, 2015). One of such non-uniform sampling approach is *data-driven sampling*, where the sampling probability of the $i$-th example depends on its norm $\|x_i\|_2$. A possible limitation of data-driven sampling is that the sampling probabilities cannot be changed during the optimization process. Recently, a method called Quartz is proposed in (Qu et al., 2015). Quartz is an SDCA-like primal-dual type stochastic optimization method whose convergence is proved with arbitrary sampling. By using arbitrary sampling scheme, it is possible to change the sampling probabilities during the optimization process.

Table 1 summarizes a selected list of recent primal-dual type stochastic optimization algorithms. Stochastic Dual Coordinate Ascent (SDCA) (Shalev-Shwartz and Zhang, 2013) is one of the most well-known non-accelerated stochastic optimization methods with uniform sampling. In the work of Iprox-SDAC (Zhao and Zhang, 2015), the authors showed that, the convergence rate can be improved by introducing the data-driven sampling probability. As mentioned above, Quartz (Qu et al., 2015) is an SDCA-like method with arbitrary sampling. On the other hand, Stochastic Primal Dual Coordinate Method (SPDC) (Zhang and Xiao, 2015) is one of the most well-known accelerated stochastic optimization method with uniform sampling. The authors in (Zhang and Xiao, 2015) also considered so-called Weighted SPDC which uses the data-driven sampling. A variant of SPDC called AdaSPDC was proposed in (Zhu and Storkey, 2015). We note that AdaSPDC is a uniform sampling-based method. The difference from the vanilla SPDC is that the step size in each iteration depends on the maximum norm of the instances selected in the dual variable update phase.

**Our contributions** As indicated in Table 1, one of our main theoretical contributions in this paper is to analyze the convergence of (Ada)SPDC when instances are sampled based on an arbitrary distribution (see §3). To the best of our knowledge, there are no existing studies analyzing the convergence of an accelerated primal-dual type stochastic optimization method with arbitrary sampling.

Our second contribution is to propose particular examples of sampling probabilities for (Ada)SPDC within the theoretical framework mentioned above (see §4). Specifically, as the first such example, we propose to periodically change the sampling probabilities based on the progress of the optimization and demonstrate its favorable empirical performance when the mini-batch size is small. We call this method as Optimality Violation-based Sampling SPDC (ovsSPDC) method. As the second example, we discuss an approach studied in AdaSDCA (Csiba et al., 2015), where sampling probabilities are changed at each iteration based again on the progress of the optimization process. The third example is based on recent safe screening studies (El Ghaoui et al., 2012; Ndiaye et al., 2015; Ogawa et al., 2013; Shibagaki et al., 2016). Safe screening allows us to identify a part of the primal and the dual variables that turned out to be zero at the optimal solution. If we can identify these variables by safe screening, we can set the sampling probabilities of those instances to zero. Similar approach is employed in a method called Affine-SDCA (Vainsencher et al., 2015).

Unfortunately, the above three particular examples of sampling probabilities have limitation in general practical use. Our third contribution in this paper is to propose two heuristic variants of ovsSPDC called ovsSPDC+ and ovsSPDC++.

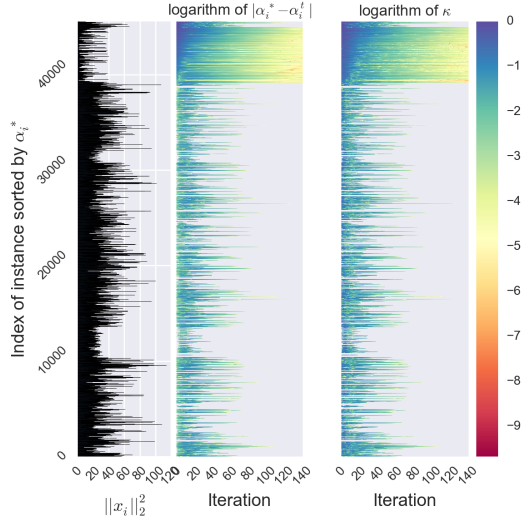

Figure 1: An illustration of optimization process when solving (1) with smoothed hinge loss, elastic net penalty, and $\lambda = 10^{-4}\|\mathrm{diag}(y)X^\top 1\|_\infty/n$ on w8a dataset by using SPDC with the uniform sampling. The left part shows the norm $\|x_i\|_2^2$ of 45546 instances sorted in the increasing order of $\alpha_i^*$ in x-axis. The center part shows how the difference between a dual variable and its optimal value $\log_{10}(1 + |\alpha_i^* - \alpha_i^t|)$ changes in the heat map representation. The right part shows the *dual optimality violation* $\log_{10}(1 + \kappa_i^t)$ (formally defined in §4). Note that the dual optimality violation is highly correlated with the values in the center part. It indicates that the dual optimality violation can be used for designing a sampling probability for non-uniform sampling.

Table 1: Selected recent primal-dual type stochastic optimization methods.

|  | Uniform | Data-driven | Arbitrary |
|---|---|---|---|
| Non Acc. | SDCA (Shalev-Shwartz and Zhang, 2013) | Iprox-SDCA (Zhao and Zhang, 2015) | Quartz (Qu et al., 2015) |
| Accelerated | SPDC, AdaSPDC (Zhang and Xiao, 2015), (Zhu and Storkey, 2015) | Weighted SPDC (Zhang and Xiao, 2015) | This paper |

Especially, the latter is useful when the data matrix $X$ is large both in the number of instances $n$ and in the number of features $d$. When we apply the heuristic variant to the ERM problem with smoothed hinge loss and elastic net penalty (meaning that both of the primal and the dual variables are sparse), the experimental results indicate that it performs almost as well as "oracle" case where one has the knowledge of the true sparsity patterns of the primal and the dual optimal solution.

Figure 1 illustrates the motivation of this study. From the center plot, we can see that about 40000 dual variables converged to the optimal values after 100 or less iterations. Obviously, those instances should not be sampled after 100 iterations, i.e., it is desirable to set the sampling probabilities of these instances to zero after 100 iterations. The left plot shows that the dual optimality violation are highly correlated with the values in the center plot. It means that, if we use the dual optimality violation for setting the sampling probabilities, we might be able to assign high sampling probabilities to instances whose corresponding dual variables are still away from the optimal values. On the other hand, the norm $\|x_i\|_2^2$ in the left plot are not highly correlated with the values in the center plot. It suggests that data-driven sampling approaches based on $\|x_i\|_2^2$ is not helpful.

In the following sections, all the proofs of theorems, lemmas, and propositions are presented in Appendix A unless otherwise stated.

## 2 Related Works

Stochastic optimization algorithms with non-uniform sampling have been studied in the past few years. Stochastic gradient descent type algorithms were studied in (Needell et al., 2014; Richtárik and Takáč, 2015; Nutini et al., 2015; Palaniappan and Bach, 2016), while accelerated coordinate descent type algorithms were studied in (Qu and Richtárik, 2016; Allen-Zhu et al., 2016). In this section, we review recent advances in primal-dual type stochastic optimization methods with non-uniform sampling.

A primal-dual type stochastic optimization algorithm alternatively iterates the primal variable update phase and the dual variable update phase. In most of the existing primal-dual type stochastic optimization algorithms, all the primal variables are updated in the primal variable update phase, while only a part of the dual variables are randomly selected and updated in the dual variable update phase.

**Definition 1** (Proper sampling probability). *Let us write the sampling probability in a dual variable update phase as $p := [p_1, \ldots, p_n]$, where $p_i$ represents the probability that the $i$-th training instance is sampled. We call $p$ is proper if $p_i > 0$ for all $i \in [n]$.*

**Definition 2** (Mini-batch size). *Let us denote $a \in [1, n]$ to be the mini-batch, i.e., the number of dual variables sampled at each of the dual variable update phase.*

As briefly discussed in §1, Table 1 summarizes a selected list of recent primal-dual type stochastic optimization algorithms. In this section, we review some of these related studies in more detail.

### 2.1 Data-driven sampling

We call a sampling probability $p$ as *data-driven distribution* if $p_i$ depends only on $\|x_i\|_2$. Iprox-SDCA (Zhao and Zhang, 2015) is a data-driven sampling variant of SDCA. In this algorithm, in the case where $f_i$ is the smoothed hinge-loss and the mini-batch size $a = 1$, a lower bound of $\mathbb{E}\left[D(\alpha^{t+1}) - D(\alpha^t)\right]$ is maximized by setting the sampling probability as $p_i \propto 1 + \|x_i\|_2^2/(\lambda n \gamma)$, and the authors showed that the convergence rate of Iprox-SDCA is faster than vanilla SDCA.

Weighted-SPDC (Zhang and Xiao, 2015) is a data-driven sampling variant of SPDC, whose sampling probability is defined as $p_i = \frac{1}{2n} + \frac{\|x_i\|_2}{2\sum_{k\in[n]} \|x_k\|_2}$. It indicates that instances whose norm $\|x_i\|_2$ are greater than others tend to be sampled more often. The authors showed that, by using the above data-driven sampling probability, the iteration complexity for obtaining an $\varepsilon$-accurate solution can be improved to $\frac{1}{n}\sum_{i\in[n]} \|x_i\|_2$ from $\max_{i\in[n]} \|x_i\|_2$ in the case of vanilla SPDC.

### 2.2 Arbitrary sampling

Quartz (Qu et al., 2015) is a dual coordinate ascent type algorithm such as SDCA. The authors analyzed the convergence property of this algorithm with a proper arbitrary sampling probability under the condition that a so-called *Expected Separable Over-approximation (ESO) property* (Richtárik and Takáč, 2015) is satisfied. In the case where $f_i$ is $1/\gamma$-smooth, $g$ is 1-strongly convex and the mini-batch size $a = 1$, the author proposed a specific sampling probability based on the norm $\|x_i\|_2^2$, which results in similar types of sampling probability as Iprox-SDCA. In §3, we will discuss more about Quartz.

**Algorithm 1:** AdaSPDC with non-uniform sampling

**Initialize**: $a$ (mini-batch size), $w^0, \alpha^0$ (initial solutions), $\bar{w}^0 \leftarrow w^0, \bar{\alpha}^0 \leftarrow \alpha^0$
**For** $t = 1, 2, \ldots$ **to** converged **do**
Generate random index from $p$ $a$ times with replacement (denote the set of random indices as $K$)
Update the dual and the primal coordinates:

$$\alpha_i^{t+1} = \begin{cases} \operatorname*{argmax}_{\beta \in \mathbb{R}} \left\{ -\beta \langle x_i, \bar{w}^t \rangle - f_i^*(-\beta) - \frac{p_i n}{2\sigma_i}(\beta - \alpha_i^t)^2 \right\} & (i \in K) \\ \alpha_i^t & (i \notin K) \end{cases}$$

$$\bar{\alpha}_i^{t+1} = \alpha_i^t + \frac{1}{ap_i n}(\alpha_i^{t+1} - \alpha_i^t) \quad (i \in [n])$$

$$w^{t+1} = \operatorname*{argmin}_{w \in \mathbb{R}^d} \left\{ \lambda g(w) - \frac{1}{n} w^\top X^\top \bar{\alpha}^{t+1} + \frac{1}{2\tau} \|w - w^t\|_2^2 \right\}$$

$$\bar{w}^{t+1} = w^{t+1} + \theta(w^{t+1} - w^t),$$

where $\theta := \max\left\{ 1 - \left(\frac{1}{2\tau\lambda}\right)^{-1}, 1 - \left(\max_{i \in [n]} \frac{1}{ap_i} + \frac{n}{2a\sigma_i \gamma}\right)^{-1} \right\}$, $\sigma_i$ and $\tau$ are the parameters that would be determined if the probability vector $p$ is determined (see Theorem 4 and Theorem 5).

## 3 SPDC with arbitrary sampling

In this section, we analyze the convergence property of AdaSDPC under the situation that dual variables are sampled from a proper arbitrary sampling probability. Algorithm 1 describes AdaSDPC with non-uniform sampling. Similar convergence analysis of vanilla SPDC (Zhang and Xiao, 2015) with a proper sampling is presented in Appendix D.

**Lemma 3.** *Assume that $g$ is 1-strongly convex and $f_i$ is convex and $1/\gamma$-smooth for each $i \in [n]$. Let $p \in (0, 1/a]^n$ be a sampling probability, $w^*$ and $\alpha^*$ are the optimal solution of (1) and (2), respectively, and define $\Delta_t$ as*

$$\Delta_t := \frac{\|w^t - w^*\|_2^2}{1/(1/2\tau + \lambda)} + \sum_{i \in [n]} \left\{ \left( \frac{1}{2\sigma_i} + \frac{\gamma}{np_i} \right) \frac{(\alpha_i^t - \alpha_i^*)^2}{a} \right\}$$
$$+ \frac{\|w^t - w^{t-1}\|_2^2}{4\tau} - \frac{(\alpha^t - \alpha^*)^\top X(w^t - w^{t-1})}{n}.$$

*Then, if the parameters $\tau, \sigma_i, \theta$ and $p$ in Algorithm 1 satisfy the following inequality:*

$$\left( \frac{1}{2a\sigma_k} - \frac{\tau \|x_k\|_2^2 \left((1 - ap_k)^2 + \theta\right)}{(ap_k n)^2} \right) \geq 0, \quad \forall k \in K$$

*then, for $t \geq 0$, Algorithm 1 achieves $\mathbb{E}[\Delta_{t+1}] \leq \theta \Delta_t$.*

If we fix the sampling probability $p$ for all $t \geq 0$, we can derive the iteration complexity of Algorithm 1 for obtaining an $\varepsilon$-accurate solution.

**Theorem 4.** *Suppose that the assumptions of Lemma 3 hold. For any proper distribution $p$ with $p_i \in (0, 1/a], \forall i \in [n]$, if we set the parameters $\tau$ and $\sigma_i$ as $\tau = \frac{aR}{2}\sqrt{\frac{\gamma}{\lambda}}, \sigma_i = \frac{np_i}{2\|x_i\|_2}\sqrt{\frac{\lambda}{\gamma}}$, where $\underline{R} := \min_{i \in [n]} \frac{p_i}{\|x_i\|_2}$, then Algorithm 1 guarantees $\mathbb{E}\left[\|w^T - w^*\|_2^2\right] \leq \varepsilon$ for $T \geq \max_{i \in [n]} \left( \frac{1}{ap_i} + \frac{\|x_i\|_2}{p_i a\sqrt{\gamma\lambda}} \right) \log \left( \frac{\Delta_0/((2\tau)^{-1} + \lambda)}{\varepsilon} \right)$ iterations.*

We can improve the iteration complexity of Theorem 4 by restricting sampling probabilities and mini-batch size $a$ as follows.

| SPDC $\tau, \sigma_i$ | Limit to $a$ | Limit to $p$ | Dominant factor of iteration complexity for obtaining $\varepsilon$-approx solution | Necessary mini-batch size for achieving optimal iteration complexity $O(\sqrt{\frac{1}{\gamma\lambda}}\log(\frac{1}{\varepsilon}))$ | |
|---|---|---|---|---|---|
| | | | | when $1/\gamma\lambda \geq n$ | $1 < 1/\gamma\lambda < n$ |
| $\tau = \frac{1}{2R}\sqrt{\frac{a\gamma}{n\lambda}}$ $\sigma_i = \frac{1}{2R}\sqrt{\frac{n\lambda}{a\gamma}}$ (Zhang and Xiao, 2015) | $\leq n$ | Uniform | $\frac{n}{a} + \frac{R\sqrt{n}}{\sqrt{a\gamma\lambda}}$ | $n$ | $n$ |
| $\tau = \frac{aR}{2}\sqrt{\frac{\gamma}{\lambda}}$ $\sigma_i = \frac{p_i n}{2\|x_i\|_2}\sqrt{\frac{\lambda}{\gamma}}$ (Theorem 4) | $\leq n$ | $p_i \in (0, \frac{1}{a}]$ | $\max_{i \in [n]}\left(\frac{1}{ap_i} + \frac{\|x_i\|_2}{ap_i\sqrt{\gamma\lambda}}\right)$ | - | - |
| | | Uniform | $\frac{n}{a} + \frac{Rn}{a\sqrt{\gamma\lambda}}$ | $n$ | $n$ |
| $\tau = \frac{aR}{2}\sqrt{\frac{n\gamma}{\lambda}}$ $\sigma_i = \frac{p_i n}{2\|x_i\|_2}\sqrt{\frac{n\lambda}{\gamma}}$ (Theorem 5) | $\leq \sqrt{n}$ | $p_i \in (0, \frac{1}{a\sqrt{n}}]$ | $\max_{i \in [n]}\left(\frac{1}{ap_i} + \frac{\|x_i\|_2}{ap_i\sqrt{n\gamma\lambda}}\right)$ | - | - |
| | | Uniform | $\frac{n}{a} + \frac{R\sqrt{n}}{a\sqrt{\gamma\lambda}}$ | $\sqrt{n}$ | Unattainable |
| $\tau = \frac{1}{2R}\sqrt{\frac{\gamma}{\lambda}}$ $\sigma_i = \frac{1}{2\|x_i\|_2}\sqrt{\frac{\lambda}{\gamma}}$ (Theorem 15) | $\geq \sqrt{n}$ | Uniform | $\frac{n}{a} + \frac{R}{\sqrt{\gamma\lambda}}$ | $\sqrt{n}$ | $n\sqrt{\gamma\lambda}$ |

Table 2: List of the dominant factor of the iteration complexity for obtaining an $\varepsilon$-approx solution and the necessary mini-batch size for achieving the optimal iteration complexity $O(\frac{1}{\sqrt{\gamma\lambda}}\log(1/\varepsilon))$ that focus on the setting of $\tau$ and $\sigma_i$, the limit to $a$, and the limit to $p$. We define $R := \max_{i \in [n]}\|x_i\|_2$.

**Theorem 5.** *Suppose that the assumptions of Lemma 3 hold and $a \leq \sqrt{n}$. For any proper distribution $p$ with $p_i \in (0, 1/(a\sqrt{n})], \forall i \in [n]$, if we set the parameters $\tau$ and $\sigma_i$ as $\tau = \frac{aR}{2}\sqrt{\frac{n\gamma}{\lambda}}$, $\sigma_i = \frac{np_i}{2\|x_i\|_2}\sqrt{\frac{n\lambda}{\gamma}}$, where $\underline{R} := \min_{i \in [n]} \frac{p_i}{\|x_i\|_2}$, then Algorithm 1 guarantees $\mathbb{E}\left[\|w^T - w^*\|_2^2\right] \leq \varepsilon$ for $T \geq \max_{i \in [n]} \left(\frac{1}{ap_i} + \frac{\|x_i\|_2}{p_i a\sqrt{n\gamma\lambda}}\right)\log\left(\frac{\Delta_0/((2\tau)^{-1}+\lambda)}{\varepsilon}\right)$ iterations.*

In the remaining part of this section, we will discuss relations of Theorems 4 and 5 with existing works. First, when we go back to uniform sampling $p_i = 1/n$ without mini-batching, the iteration complexity in Theorem 5 is reduced to that of vanilla SPDC (cite) as in the following remark.

**Remark 6.** *If $p$ is uniform and $a = 1$, then the following equality holds $\max_{i \in [n]}\left(\frac{1}{p_i} + \frac{\|x_i\|_2}{p_i\sqrt{n\lambda\gamma}}\right) = n + \sqrt{\frac{n}{\lambda\gamma}}\max_{i \in [n]}\|x_i\|_2$, where the right hand side is the dominant factor of the iteration complexity for obtaining $\varepsilon$-accurate primal solution in vanilla SPDC. It means that Theorem 5 is considered as a generalization of the result in (Zhang and Xiao, 2015).*

Next, when we consider uniform sampling $p_i = 1/n$ with mini-batching, i.e., $a > 1$, our result in Theorem 5 improves the analysis of vanilla SPDC (cite) as in the following remark.

**Remark 7.** *If $p$ is uniform and $a > 1$, we can see that Theorem 5 improves the iteration complexity in Corollary 1 in (Zhang and Xiao, 2015) from $\frac{n}{a} + \frac{\max_{i \in [n]}\|x_i\|_2\sqrt{n}}{\sqrt{a\gamma\lambda}}$ to $\frac{n}{a} + \frac{\max_{i \in [n]}\|x_i\|_2\sqrt{n}}{a\sqrt{\gamma\lambda}}$ by restricting mini-batch size $a \leq \sqrt{n}$. The previous SPDC (Zhang and Xiao, 2015) suffers form poor mini-batch efficiency and whose the total computational cost for obtaining an $\varepsilon$-approximate solution increases by using mini-batching as the authors in (Qu et al., 2015; Murata and Suzuki, 2017) described. However, Theorem 5 shows that the order of the total computational cost does not change for $a \leq \sqrt{n}$ (see Table 3). We note that the iteration complexity in Theorem 5 achieves the optimal iteration complexity of the first order algorithm $O(\frac{1}{\sqrt{\gamma\lambda}}\log(\frac{1}{\varepsilon}))$ when mini-batch size $a$ is $\sqrt{n}$ and $1/\gamma\lambda \leq n$, although the*

| Algorithm (with the uniform sampling) | Total computational cost for obtaining $\varepsilon$-approx solution when mini-batch size $a > 1$ | Necessary mini-batch size for achieving optimal iteration complexity $O(\sqrt{\frac{1}{\gamma\lambda}}\log(\frac{1}{\varepsilon}))$ | |
|---|---|---|---|
| | | when $1/\gamma\lambda \geq n$ | when $1 < 1/\gamma\lambda < n$ |
| SPDC (Zhang and Xiao, 2015) | $O\left(d\left(n + \sqrt{\frac{an}{\lambda\gamma}}\right)\log(\frac{1}{\varepsilon})\right)$ | $n$ | $n$ |
| APCG (Lin et al., 2014) | $O\left(d\left(n + \sqrt{\frac{an}{\lambda\gamma}}\right)\log(\frac{1}{\varepsilon})\right)$ | $n$ | $n$ |
| Katyusha (Allen-Zhu, 2016) | $O\left(d\left(n + \sqrt{\frac{an}{\lambda\gamma}}\right)\log(\frac{1}{\varepsilon})\right)$ | $n$ | $n$ |
| DASVRDA (Murata and Suzuki, 2017) | $O\left(d\left(n + \sqrt{\frac{n}{\lambda\gamma}} + \frac{a}{\sqrt{\lambda\gamma}}\right)\log(\frac{1}{\varepsilon})\right)$ | $\sqrt{n}$ | $n\sqrt{\lambda\gamma}$ |
| SPDC (Theorem 5 and 15) | $O\left(d\left(n + \sqrt{\frac{n}{\lambda\gamma}}\right)\log(\frac{1}{\varepsilon})\right)$ (for $a \leq \sqrt{n}$) $O\left(d\left(n + \frac{a}{\sqrt{\lambda\gamma}}\right)\log(\frac{1}{\varepsilon})\right)$ (for $a \geq \sqrt{n}$) | $\sqrt{n}$ $\sqrt{n}$ | Unattainable $n\sqrt{\lambda\gamma}$ |

Table 3: List of the total computational cost for obtaining an $\varepsilon$-approx solution and the necessary iteration complexity for obtaining the optimal iteration complexity $O(\sqrt{\frac{1}{\gamma\lambda}}\log(\frac{1}{\varepsilon}))$ on each algorithm.

*iteration complexity in (Zhang and Xiao, 2015) needs mini-batch size $a = n$. We can also improve the results in (Zhang and Xiao, 2015) for $a \geq \sqrt{n}$ (see Appendix A and B for more detail).*

Tables 2 and 3 compares the convergence properties of our analysis with existing analysis. See also Appendix B for more detailed analysis and experiments for SPDC with mini-batching.

**Comparison with Weighted SPDC**  We compare our analysis with the theoretical properties of Weighted SPDC.

**Remark 8.** *If $\max_{i\in[n]}\|x_i\|_2 \leq \frac{1}{\sqrt{n}}(\sum_{k\in[n]}\|x_k\|_2) + \sqrt{\lambda\gamma}(n - \sqrt{n})$ holds and we set $p$ with $p_i \propto \|x_i\|_2 + \sqrt{\lambda\gamma n}$ and $a = 1$, then the $p$ minimizes the dominant factor of the iteration complexity $\max_{i\in[n]}\frac{1}{p_i} + \frac{n}{2\gamma\sigma_i}$ in Theorem 5 and makes the iteration complexity one-half beside Weighted SPDC (see Appendix C for more detail).*

**Comparison with Quartz**  We compare our analysis with the theoretical properties of Quartz.

**Remark 9.** *Theorem 4 shows that the dominant term of the iteration complexity depends on $1/\sqrt{\lambda\gamma}$. Compared with the corresponding dominant term $1/\lambda\gamma$ in Quartz, our proposed method provides the accelerated convergence rate as indicated in Table 1.*

**Remark 10.** *When the mini-batch size $a > 1$, the ESO property in Quartz is restrictive in the sense that it is hard to find proper sampling probability except the uniform one. Therefore, the authors in (Qu et al., 2015) consider non-uniform sampling probability only when $a = 1$. On the other hand, as Theorem 4 indicates, it is possible to find proper non-uniform sampling probabilities even when $a > 1$ as long as $p_i \in (0, 1/a], \forall i \in [n]$. Actually, we present a particular example of such a proper sampling probability for $a \geq 1$ in § 4.1.*

## 4 Examples of Sampling Probabilities

In this section, we propose three examples of sampling probabilities for SPDC within the theoretical framework developed in §3.

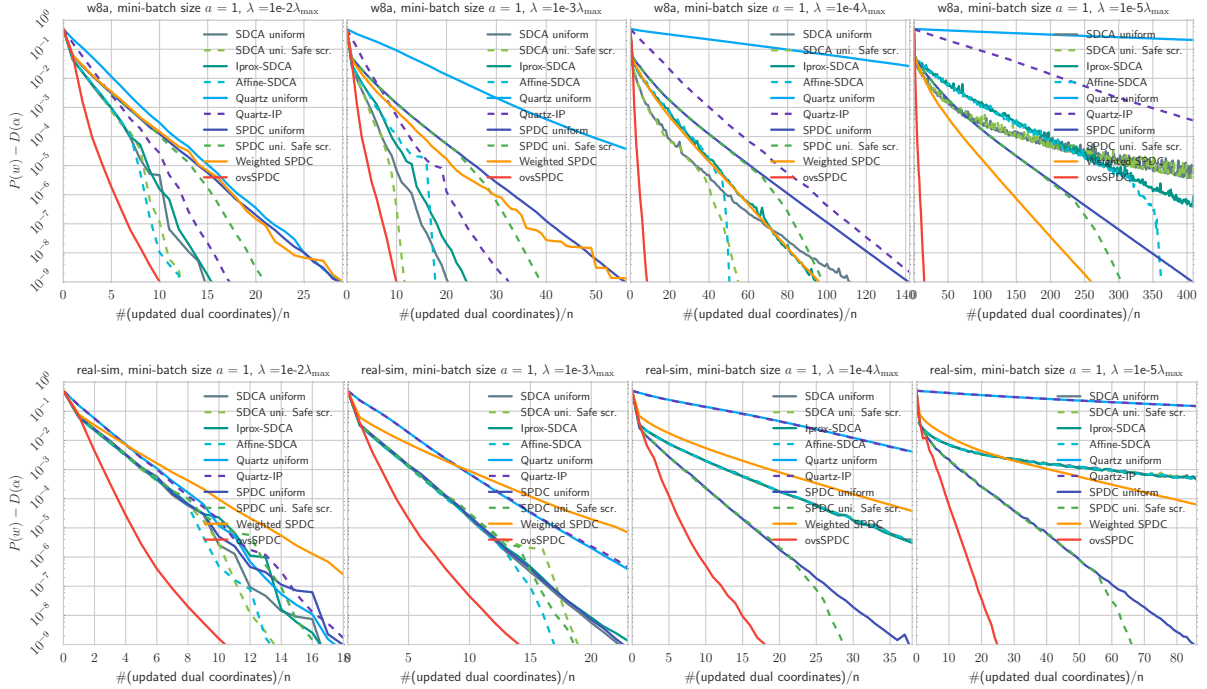

Figure 2: The illustration of the convergence speed of each method on `w8a` (not normalized) and `real-sim` (normalized).

### 4.1 Proper sampling based on optimality violations

Here, we propose a proper sampling probability based on optimality violations at the current solution. The basic idea is to assign larger weights to the instances whose corresponding dual variables are still away from the optimal values. Let

$$\kappa_i^t := |-\alpha_i^t - \nabla f_i(x_i^\top w^t)|, i \in [n]. \tag{3}$$

This quantity, which we call *dual optimality violation*, implies the violation of the optimality condition of the current primal and dual variables (see Figure 1). By using $\kappa_i^t$, we set

$$\forall i \in [n], \ p_i = \begin{cases} \frac{\rho_i \|x_i\|_2}{\sum_{k \in [n]} \rho_k \|x_k\|_2} & (\overline{p} \leq 1/(a\sqrt{n})), \\ \frac{1-\zeta}{n} + \zeta \frac{\rho_i \|x_i\|_2}{\sum_{k \in [n]} \rho_k \|x_k\|_2} & (\text{otherwise}), \end{cases}$$

where $\rho_i := \kappa_i^t + \min_{k \in \{l \in [n] | \kappa_l^t \neq 0\}} \kappa_k^t$, $\overline{p} := \max_{i \in [n]} \rho_i \|x_i\|_2 / (\sum_{k \in [n]} \rho_k \|x_k\|_2)$, $\zeta := (1/(a\sqrt{n}) - 1/n)/(\overline{p} - 1/n)$. Here, it is important to note that $p$ satisfy the conditions in Theorem 5, i.e., $p_i \in (0, 1/(a\sqrt{n})] \ \forall i \in [n]$. This is possible because we define all the $\rho_i$ to be nonzero by adding an offset term $\min_{k \in \{l \in [n] | \kappa_l^t \neq 0\}} \kappa_k^t$.

We calculate $\kappa^t$ and update $p$ every $\lceil n/a \rceil$ iterations. We first need to compute $\kappa_i^t$ for all $i \in [n]$, which takes $O(nd)$ computation. Furthermore, for non-uniform sampling, we use Alias method (Walker, 1977) which takes $O(n)$ for preprocessing time, and $O(1)$ for sampling. Thus, the additional cost for non-uniform sampling is $O(nd) + O(n)$ per $\lceil n/a \rceil$ iterations, meaning that the order of the computational time is kept same as uniform sampling.

## 4.2 Non proper sampling based on optimality violation

In the previous section, $p_i$ is nonzero even the optimality violation $\kappa_i^t$ is zero. If we do so, the sampling probability $p$ would not be proper anymore, and the theoretical results discussed in §3 cannot be used for proving the convergence.

In a restricted case that the loss function is the squared loss and $a = 1$, the authors in (Csiba et al., 2015) showed the convergence of AdaSDCA even if $p_i$ is exactly zero when the optimality violation $\kappa_i^t$ is zero as long as $\kappa^t$ is updated at each iteration. This is possible because $\kappa_i^t = 0$ means $\alpha_i^{t+1} = \alpha_i^t$. We can exploit this fact also for SPDC. For example, in smoothed hinge loss, the following proposition holds.

**Proposition 11.** *At the $t$-th iteration of Algorithm 1, if $f_i$ is smoothed hinge loss and $\gamma - p_i n/\sigma_i \neq 0$ then $|-\alpha_i^t - \nabla f_i(x_i^\top \bar{w}^t)| = 0 \Rightarrow \alpha_i^{t+1} = \alpha_i^t$.*

By using this proposition and computing $\kappa^t$ at each iteration, we can prove the convergence even if $p_i = 0$ for $-\alpha_i^t - \nabla f_i(x_i^\top \bar{w}^t) = 0$.

We can also use the primal optimality violation defined as

$$\psi_j^t := \left| w_j^t - \nabla g_j^* \left( \frac{1}{\lambda n} X_{:j}^\top \alpha^t \right) \right|. \quad (4)$$

For example, in elastic net penalty case the following proposition holds.

**Proposition 12.** *At the $t$-th iteration of Algorithm 1, if $g$ is elastic net penalty, $\lambda + 1/\tau \neq 0$, and $\text{sign}(w_j^{t+1}) = \text{sign}(w_j^t)$, then $|w_j^t - \nabla g_j^* \left( \frac{1}{\lambda n} X_{:j}^\top \bar{\alpha}^{t+1} \right)| = 0 \Rightarrow w_j^{t+1} = w_j^t$.*

This proposition implies that we can prove convergence even if we skip the update of $w_j^t$ if $w_j^t - \nabla g_j^* \left( \frac{1}{\lambda n} X_{:j}^\top \bar{\alpha}^{t+1} \right) = 0$, $\lambda + 1/\tau \neq 0$ and $\text{sign}(w_j^{t+1}) = \text{sign}(w_j^t)$ in the primal variable update phase as long as $\psi^t$ is updated at each iteration.

## 4.3 Non proper sampling based on safe screening

Another approach for using non-proper sampling probability within the theoretical framework in §3 is to introduce an idea in safe instance screening (Ogawa et al., 2013). Safe instance screening allows us to identify a part of non-active dual variables (i.e., $\alpha_i^* = 0$) without knowing the optimal solution itself. We can prove convergence even if $p_i = 0$ if the corresponding dual variable $\alpha_i$ is screened-out by safe instance screening. We can set $p_i > 0$ as an arbitrary value for the other $i$.

Affine-SDCA (Vainsencher et al., 2015) exploits this idea, i.e., it changes $p$ during the optimization process by setting $p_i = 0$ if the corresponding dual variable is identified as non-active by safe instance screening (see Theorem 10 in (Vainsencher et al., 2015)). We can use the same idea for SPDC as stated in the following proposition.

**Proposition 13.** *Suppose that the assumptions in Lemma 3 hold, $a = 1$ and the initial solutions satisfy $\Delta_0 \leq \varepsilon'$. Let $\mathcal{I}$ be the set of indices which are identified as non-active dual variables by safe instance screening. If we set $p_i = 0$ for $i \in \mathcal{I}$ and arbitrary value in $(0, 1/\sqrt{n'}]$ otherwise, $\tau = \frac{aR}{2}\sqrt{\frac{n'\gamma}{\lambda}}$, $\sigma_i = \frac{n'p_i}{2\|x_i\|_2}\sqrt{\frac{n'\lambda}{\gamma}}$, where $n' := n - |\mathcal{I}|$, $\underline{R} := \min_{i \in [n] \setminus \mathcal{I}} p_i / \|x_i\|_2$, $\underline{p} := \min_{i \in [n] \setminus \mathcal{I}} p_i$, then Algorithm 1 guarantees, for $T \geq \max_{i \in [n] \setminus \mathcal{I}} \left( \frac{1}{ap_i} + \frac{\|x_i\|_2}{p_i a \sqrt{n'\gamma\lambda}} \right) \log \left( \frac{\varepsilon'}{\varepsilon} \right)$, $\mathbb{E}[\Delta_T] \leq \varepsilon$.*

There are also several studies on safe feature screening which allows us to identify a part of non-active primal variables (i.e., $w_j^* = 0$) (El Ghaoui et al., 2012; Ndiaye et al., 2015). We can also prove the convergence even if we skip the update of $w_j^t$ if it is identified as non-active by safe feature screening.

## 4.4 Experiments

Here, we demonstrate the effectiveness of the method proposed in §4.1. through numerical experiments on ERM (1) with smoothed hinge loss and elastic net penalty. We set $\gamma = 1$ and considered four different choices of the regularization parameter $\lambda \in \{10^{-2}, 10^{-3}, 10^{-4}, 10^{-5}\}\lambda_{\max}$, where $\lambda_{\max} := \|\text{diag}(y)X1\|_\infty/n$. Table 4 shows the datasets used in the experiments. Those datasets were obtained form LIBSVM Data (Chang and Lin, 2011). We set $a = 1$ because the competing methods with non-uniform sampling do not support mini-batching ($a > 1$). Due to the space limitation, we only provide the experimental results on `w8a` and `real-sim`. The experimental results on the other datasets are presented in Appendix F. We implemented all the algorithms using C++ and Eigen, and attached our code as a supplementary material. It will be published on the web after the paper is accepted. All the computations were performed on Intel Xeon CPU E5-2687W v3 (3.10GHz), 256GB RAM.

Table 4: Benchmark datasets used in the experiments.

| Dataset name | Normalized | $n$ | $d$ | #(nnz)/$nd$ |
|---|---|---|---|---|
| w8a |  | 45,546 | 300 | 0.042418 |
| ijcnn1-train |  | 35,000 | 22 | 0.590909 |
| a9a |  | 32,561 | 123 | 0.112757 |
| real-sim | ✓ | 72,201 | 20,958 | 0.002451 |
| rcv1-test | ✓ | 677,399 | 47,236 | 0.001548 |
| rcv1-train | ✓ | 20,242 | 47,236 | 0.001567 |

#(nnz) indicates the number of non-zero elements. "Normalized" indicates $\forall i \in [n], \|x_i\|_2 = 1$, meaning that data-driven sampling is reduced to the uniform sampling.

In Figure 5, we compared the proposed method (denoted as osvSPDC) with nine algorithms. Even though `w8a` is not normalized (see Figure 1), Iprox-SDCA is not effective. Similar results were also reported by (Vainsencher et al., 2015). The results show that the proposed method is faster than the other methods in all cases. In particular, it is substantially faster when $\lambda$ is small.

**Algorithm 2: ovsSPDC+**

**input** : $a$ (mini-batch size), $w^0, \alpha^0$

**for** $t = 1, 2, \ldots$ **to** *converged* **do**
$\quad \alpha_i^t \leftarrow \text{update}(\alpha_i^{t-1}) \ \forall i \in [n]$
$\quad w^t \leftarrow \text{update}(w^{t-1})$
$\quad$ Compute $\kappa^t$ and set $p$ as (5)
$\quad (\hat{w}^0, \hat{\alpha}^0) \leftarrow (w^t, \alpha^t)$
$\quad$ **for** $u = 1, 2, \ldots \lceil n/a \rceil$ **do**
$\quad\quad$ Randomly pick up a subset indicies $K$
$\quad\quad \hat{\alpha}_i^u \leftarrow \text{update}(\hat{\alpha}_i^{u-1}) \ \forall i \in K$
$\quad\quad \hat{w}^u \leftarrow \text{update}(\hat{w}^{u-1})$
$\quad$ **if**
$\quad P_\lambda(\hat{w}^u) - D_\lambda(\hat{\alpha}^u) < P_\lambda(w^{(t)}) - D_\lambda(\alpha^{(t)})$
$\quad$ **then**
$\quad\quad (w^t, \alpha^t) \leftarrow (\hat{w}^u, \hat{\alpha}^u)$

**Algorithm 3: ovsSPDC++**

**input** : $a$ (mini-batch size), $w^0, \alpha^0$

**for** $t = 1, 2, \ldots$ **to** *converged* **do**
$\quad \alpha_i^t \leftarrow \text{update}(\alpha_i^{t-1}) \ \forall i \in [n]$
$\quad w_j^t \leftarrow \text{update}(w_j^{t-1}) \ \forall j \in [d]$
$\quad$ Compute $\kappa^t, \psi^t$ and set $p$ as (5)
$\quad (\hat{w}^0, \hat{\alpha}^0) \leftarrow (w^t, \alpha^t)$
$\quad$ **for** $u = 1, 2, \ldots \lceil n/a \rceil$ **do**
$\quad\quad$ Randomly pick up a subset indices $K$
$\quad\quad \hat{\alpha}_i^u \leftarrow \text{update}(\hat{\alpha}_i^{u-1}) \ \forall i \in K$
$\quad\quad \hat{w}_j^u \leftarrow \text{update}(\hat{w}_j^{u-1}) \ \forall j \in \{m \in [d] | \psi_m^t \neq 0\}$
$\quad$ **if**
$\quad P_\lambda(\hat{w}^u) - D_\lambda(\hat{\alpha}^u) < P_\lambda(w^{(t)}) - D_\lambda(\alpha^{(t)})$
$\quad$ **then**
$\quad\quad (w^t, \alpha^t) \leftarrow (\hat{w}^u, \hat{\alpha}^u)$

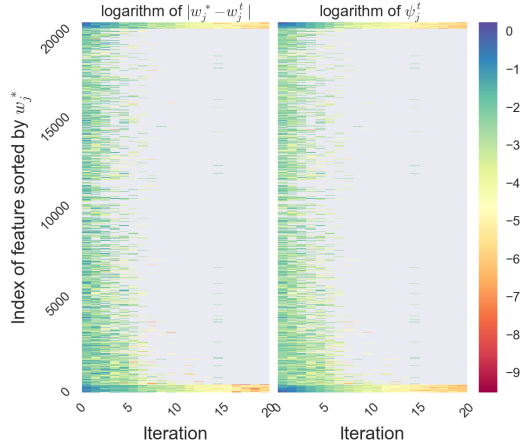

Figure 3: An illustration of optimization process when solving (1) with smoothed hinge loss, elastic net penalty, and $\lambda = 10^{-2}\lambda_{\max}$ on `real-sim` dataset by using SPDC with the uniform sampling in the same manner as Figure 1. We observe that the primal optimality violation is also highly correlated with the difference between a primal variable and its optimal value, suggesting that the primal optimality violation can be also used for selectively updating the primal variables.

## 5 Efficient heuristic variants

Unfortunately, each of the three particular examples of sampling probabilities discussed in §4 have limitation in general practical use. ovsSPDC presented in §4.1 works well when the mini-batch size $a$ is small as demonstrated in the experiments in §4.4. However, when $a$ is large, performance of ovsSPDC is not satisfactory. The method described in §4.2 is impractical as mentioned in (Csiba et al., 2015) about AdaSDCA. Specifically, in this method (and also in AdaSDCA), one must compute $\kappa^t$ and $\psi^t$ with the computational cost in $O(nd)$ at each iteration. Finally, the method based on safe screening in §4.3 is not so effective especially in the early stage of the optimization because it is too conservative as we observed in the experiments in §4.4.

In this section, in order to overcome these limitations, we go beyond the theoretical framework developed in §3, propose two heuristic variants of ovsSPDC. Algorithms 2 and 3 describe these two heuristic variants, each of which is called ovsSPDC+ and ovsSPDC++, respectively. ovsSPDC++, we assume the separability of the penalty function (i.e., $g(w) = \sum_{j \in [d]} g_j(w_j)$). ovsSPDC+ is particularly designed for the case where the number of instances $n$ is large, and the dual optimal variables are sparse (as is the case of hinge loss or its variants). On the other hand, ovsSPDC++ is particularly designed for the case where both the number of instances $n$ and the number of features $d$ are large, and both of the primal and the dual optimal variables are sparse. (as is the case of hinge loss or its variants + sparsity inducing penalty such as elastic net).

ovsSPDC+ has outer and inner loops. In the outer loop, all the primal and the dual variable are updated for guaranteeing the convergence, and the optimality violation $\kappa^t$ is computed for setting the sampling probability as

$$p_i = \begin{cases} 1/|\{i \in [n] \mid \kappa_i^t \neq 0\}| & (\kappa_i \neq 0), \\ 0 & (\kappa_i = 0), \end{cases} \quad (5)$$

which means that dual variables satisfying the optimality conditions are never updated. In the inner loop, all the primal variables and a part of the dual variables as specified by the sampling probability in (5) are updated [*].

---

[*] In the inner loop, the number of dual variables are considered to be $|\{i \in [n] \mid \kappa_i^t \neq 0\}|$ instead of $n$.

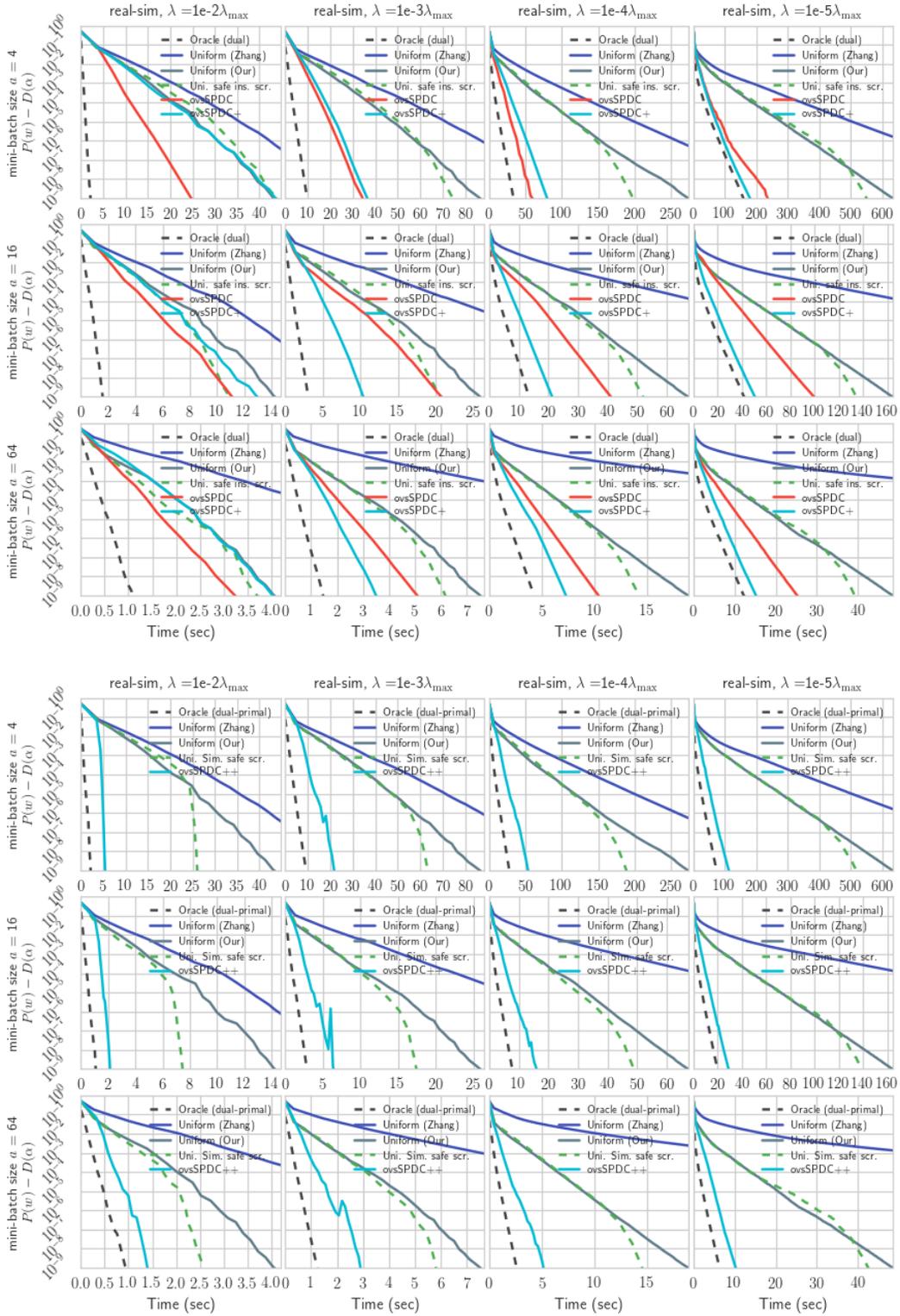

Figure 4: The illustrations of the convergence speed of each method on `real-sim`.

ovsSPDC++, we also consider updating only a part of the primal variables in its inner loop. To do this, it computes $\psi^t$ in the outer loop, and only updates the primal variables whose $\psi^t_j \neq 0$. Figure 3 shows $|w^*_j - w^t_j|$ and $\psi^t_j$ in the process of the optimization. Since $|w^*_j - w^t_j|$ and $\psi^t_j$ are highly correlated, $\psi^t_j$ can be used as a criterion for selecting the primal variables to be updated in the inner loop as is done in ovsSDPC++.

### 5.1 Experiments

We demonstrate the advantage of the proposed heuristic variants through numerical experiments. The experimental setups are the same as those in §4.4.

We first compare the computational time of ovsSPDC+ (Algorithm 2), SPDC with the uniform sampling (Zhang and Xiao, 2015), SPDC with the uniform sampling (Theorem 5), SPDC using safe instance screening, and *Oracle (dual)*. Here, Oracle (dual) means that the sparsity pattern of $\alpha^*_i$ has been already known, meaning that the optimization is carried out only for those who are active. We considered two different choices of mini-batch size $a \in \{4, 64\}$. Figure 4 shows the result on real-sim. ovsSPDC are comparable ovsSPDC+ is still effective in the case of $a = 4$, although it is less effective than ovsSPDC+ when $a = 64$ and $\lambda$ is small.

Next, we compared the computational time of ovsSPDC++ (Algorithm 3) , SPDC with the uniform sampling, SPDC using simultaneous safe screening (Shibagaki et al., 2016), and Oracle (dual-primal). Again, Oracle (dual-primal) means that the sparsity patterns of $\alpha^*_i$ and $w^*_j$ have been already known, meaning that the optimization is carried out only for those who are active. Figure 4 shows the result on real-sim, indicating that ovsSPDC++ is comparable to Oracle (dual-primal).

# Acknowledgements

IT was partially supported by JST CREST 15656320, MEXT KAKENHI 16H06538, and RIKEN Center for Advanced Intelligence Project.

# A Proofs

In this Appendix, we prove Lemma 3, Theorem 4, 5, 15, and Proposition 11, 12.

## A.1 Proof of Lemma 3

We extended the proof in (Zhang and Xiao, 2015).

**Lemma 3.** *Assume that $g$ is 1-strongly convex and $f_i$ is convex and $1/\gamma$-smooth for each $i \in [n]$. Let $p \in (0, 1/a]^n$ be a probability vector, $w^*$ and $\alpha^*$ are the optimal solution of (1) and (2), respectively, and define $\Delta_t$ as*

$$\Delta_t := \left(\frac{1}{2\tau} + \lambda\right) \|w^t - w^*\|_2^2 + \sum_{i \in [n]} \left\{ \left(\frac{1}{2\sigma_i} + \frac{\gamma}{np_i}\right) \frac{(\alpha_i^t - \alpha_i^*)^2}{a} \right\} + \frac{\|w^t - w^{t-1}\|_2^2}{4\tau} - \frac{(\alpha^t - \alpha^*)^\top X(w^t - w^{t-1})}{n}.$$

*Then, if the parameters $\tau, \sigma_i, \theta$ and $p$ in Algorithm 1 satisfy the following inequality:*

$$\left(\frac{1}{2a\sigma_k} - \frac{\tau \|x_k\|_2^2 ((1-ap_k)^2 + \theta)}{(ap_k n)^2}\right) \geq 0, \quad \forall k \in K \tag{6}$$

*then, for $t \geq 0$, Algorithm 1 achieves $\mathbb{E}[\Delta_{t+1}] \leq \theta \Delta_t$.*

*Proof.* Let $\tilde{\alpha}_i$ as follows:

$$\tilde{\alpha}_i := \underset{\beta \in \mathbb{R}}{\operatorname{argmin}} \left\{ \beta \langle x_i, \bar{w}^t \rangle + f_i^*(-\beta) + \frac{p_i n}{2\sigma_i}(\beta - \alpha_i^t)^2 \right\}.$$

From $1/\gamma$-smoothness of $f_i$, the objective function of the above optimization problem is $(\gamma + \frac{p_i n}{\sigma_i})$-strongly convex. Therefore, we have the following inequality:

$$(\forall i \in [n]) \ \alpha_i^* \langle x_i, \bar{w}^t \rangle + f_i^*(-\alpha_i^*) + \frac{p_i n}{2\sigma_i}(\alpha_i^* - \alpha_i^t)^2$$
$$\geq \tilde{\alpha}_i \langle x_i, \bar{w}^t \rangle + f_i^*(-\tilde{\alpha}_i) + \frac{p_i n}{2\sigma_i}(\tilde{\alpha}_i - \alpha_i^t)^2 + \left(\frac{\gamma}{2} + \frac{p_i n}{2\sigma_i}\right)(\tilde{\alpha}_i - \alpha_i^*)^2. \tag{7}$$

Since $f_i^*$ is $\gamma$-strongly convex, we have

$$f_i^*(-\tilde{\alpha}_i) + \tilde{\alpha}_i \langle x_i, w^* \rangle \geq f_i^*(-\alpha_i^*) + \alpha_i^* \langle x_i, w^* \rangle + \frac{\gamma}{2}(\tilde{\alpha}_i - \alpha_i^*)^2. \tag{8}$$

By combining (7) and (8), we have

$$\frac{p_i n}{2\sigma_i}(\alpha_i^t - \alpha_i^*)^2 \geq \left(\frac{p_i n}{2\sigma_i} + \gamma\right)(\tilde{\alpha}_i - \alpha_i^*)^2 + \frac{p_i n}{2\sigma_i}(\tilde{\alpha}_i - \alpha_i^t)^2 + (\tilde{\alpha}_i - \alpha_i^*)\langle x_i, \bar{w}^t - w^* \rangle.$$

Since

$$\mathbb{E}\left[(\alpha_i^{t+1} - \alpha_i^*)^2\right] = ap_i(\tilde{\alpha}_i - \alpha_i^*)^2 + (1 - ap_i)(\alpha_i^t - \alpha_i^*)^2$$
$$\mathbb{E}\left[(\alpha_i^{t+1} - \alpha_i^t)^2\right] = ap_i(\tilde{\alpha}_i - \alpha_i^t)^2$$
$$\mathbb{E}\left[\alpha_i^{t+1}\right] = ap_i \tilde{\alpha}_i + (1 - ap_i)\alpha_i^t,$$

we have

$$\frac{p_i n}{2\sigma_i}(\alpha_i^t - \alpha_i^*)^2 \geq \left(\frac{n}{2a\sigma_i} + \frac{\gamma}{ap_i}\right) \mathbb{E}\left[(\alpha_i^{t+1} - \alpha_i^*)^2\right] - (1 - ap_i)\left(\frac{n}{2a\sigma_i} + \frac{\gamma}{ap_i}\right)(\alpha_i^t - \alpha_i^*)^2$$
$$+ \frac{n}{2a\sigma_i}\mathbb{E}\left[(\alpha_i^{t+1} - \alpha_i^t)^2\right] + \left(\frac{1}{ap_i}\mathbb{E}\left[\alpha_i^{t+1} - \alpha_i^*\right] + \alpha_i^t - \alpha_i^*\right)\langle x_i, \bar{w}^t - w^*\rangle.$$

By summing $i = 1, \ldots, n$ and dividing by n, we have

$$\sum_{i\in[n]}\left\{\left(\frac{1}{2\sigma_i} + \frac{(1-ap_i)\gamma}{p_i n}\right)\frac{(\alpha_i^t - \alpha_i^*)^2}{a}\right\} \geq \sum_{i\in[n]}\left\{\left(\frac{1}{2\sigma_i} + \frac{\gamma}{p_i n}\right)\frac{\mathbb{E}\left[(\alpha_i^{t+1} - \alpha_i^*)^2\right]}{a}\right\} + \frac{\mathbb{E}\left[\sum_{k\in K}(\alpha_k^{t+1} - \alpha_k^t)^2\right]}{2a\sigma_i}$$
$$+ \mathbb{E}\left[\left\langle u^t - u^* + \sum_{k\in K}\left\{\frac{1}{ap_k n}(\alpha_k^{t+1} - \alpha_k^t)x_k\right\}, \bar{w}^t - w^*\right\rangle\right], \quad (9)$$

where $u^t := (1/n)\sum_{i\in[n]}\alpha_i x_i$, $u^* := (1/n)\sum_{i\in[n]}\alpha_i^* x_i$.

Let $w^{t+1}$ as follows:

$$w^{t+1} := \operatorname*{argmin}_{w\in\mathbb{R}^d}\left\{\lambda g(w) - \left\langle u^t + \sum_{k\in K}\left\{\frac{1}{ap_k n}(\alpha_k^{t+1} - \alpha_k^t)x_k\right\}, w\right\rangle + \frac{1}{2\tau}\|w - w^t\|_2^2\right\}.$$

From 1-strongly convexity of $g$, the objective function of the above optimization problem is $(\frac{1}{\tau} + \lambda)$-strongly convex. Therefore, we have

$$\lambda g(w^*) - \left\langle u^t + \sum_{k\in K}\left\{\frac{1}{ap_k n}(\alpha_k^{t+1} - \alpha_k^t)x_k\right\}, w^*\right\rangle + \frac{1}{2\tau}\|w^t - w^*\|_2^2 \geq$$
$$\lambda g(w^{t+1}) - \left\langle u^t + \sum_{k\in K}\left\{\frac{1}{ap_k n}(\alpha_k^{t+1} - \alpha_k^t)x_k\right\}, w^{t+1}\right\rangle + \frac{1}{2\tau}\|w^t - w^{t+1}\|_2^2 + \left(\frac{1}{2\tau} + \frac{\lambda}{2}\right)\|w^{t+1} - w^*\|_2^2.$$
(10)

Since $g$ is 1-strongly convex, we have

$$\langle u^*, w^{t+1}\rangle + \lambda g(w^{t+1}) \geq \langle u^*, w^*\rangle + \lambda g(w^*) + \frac{\lambda}{2}\|w^{t+1} - w^*\|_2^2. \quad (11)$$

By combining (10) and (11), and taking expectation, we have

$$\frac{\|w^t - w^*\|_2^2}{2\tau} \geq \left(\frac{1}{2\tau} + \lambda\right)\mathbb{E}\left[\|w^{t+1} - w^*\|_2^2\right] + \frac{\mathbb{E}\left[\|w^{t+1} - w^t\|_2^2\right]}{2\tau}$$
$$- \mathbb{E}\left[\left\langle u^t - u^* + \sum_{k\in K}\left\{\frac{1}{ap_k n}(\alpha_k^{t+1} - \alpha_k^t)x_k\right\}, w^{t+1} - w^*\right\rangle\right]. \quad (12)$$

By combining (9) and (12), we have

$$\frac{\|w^t - w^*\|_2^2}{2\tau} + \sum_{i\in[n]}\left\{\left(\frac{1}{2\sigma_i} + \frac{(1-ap_i)\gamma}{p_i n}\right)\frac{(\alpha_i^t - \alpha_i^*)^2}{a}\right\} - \theta\frac{(\alpha^t - \alpha^*)^\top X(w^t - w^{t-1})}{n}$$
$$\geq \left(\frac{1}{2\tau} + \lambda\right)\mathbb{E}\left[\|w^{t+1} - w^*\|_2^2\right] + \sum_{i\in[n]}\left\{\left(\frac{1}{2\sigma_i} + \frac{\gamma}{p_i n}\right)\frac{\mathbb{E}\left[(\alpha_i^{t+1} - \alpha_i^*)^2\right]}{a}\right\} + \frac{\mathbb{E}\left[\sum_{k\in K}(\alpha_k^{t+1} - \alpha_k^t)^2\right]}{2a\sigma_i}$$
$$+ \frac{\mathbb{E}\left[\|w^{t+1} - w^t\|_2^2\right]}{2\tau} - \mathbb{E}\left[\underbrace{\left\langle u^t - u^* + \sum_{k\in K}\left\{\frac{1}{ap_k n}(\alpha_k^{t+1} - \alpha_k^t)x_k\right\}, w^{t+1} - w^t - \theta(w^t - w^{t-1})\right\rangle}_{v}\right]. \quad (13)$$

By Letting $\delta := [(ap_1 n)^{-1}, \cdots, (ap_n n)^{-1}] \in \mathbb{R}^n$, and be $\odot$ as the element-wise product operator, we have the following equality:

$$v = \frac{(\alpha^t - \alpha^*)^\top X(w^{t+1} - w^t)}{n} - \theta\frac{(\alpha^t - \alpha^*)^\top X(w^t - w^{t-1})}{n}$$
$$+ (\delta \odot (\alpha^{t+1} - \alpha^t))^\top X(w^{t+1} - w^t) - \theta(\delta \odot (\alpha^{t+1} - \alpha^t))^\top X(w^t - w^{t-1})$$
$$= \frac{(\alpha^{t+1} - \alpha^*)^\top X(w^{t+1} - w^t)}{n} - \frac{(\alpha^{t+1} - \alpha^t)^\top X(w^{t+1} - w^t)}{n} - \theta\frac{(\alpha^t - \alpha^*)^\top X(w^t - w^{t-1})}{n}$$
$$+ (\delta \odot (\alpha^{t+1} - \alpha^t))^\top X(w^{t+1} - w^t) - \theta(\delta \odot (\alpha^{t+1} - \alpha^t))^\top X(w^t - w^{t-1})$$
$$= \frac{(\alpha^{t+1} - \alpha^*)^\top X(w^{t+1} - w^t)}{n} - \theta\frac{(\alpha^t - \alpha^*)^\top X(w^t - w^{t-1})}{n}$$
$$+ \sum_{k\in K}\left\{\delta_k(1 - ap_k)(\alpha_k^{t+1} - \alpha_k^t)\langle x_k, (w^{t+1} - w^t)\rangle\right\} - \theta\sum_{k\in K}\left\{\delta_k(\alpha_k^{t+1} - \alpha_k^t)\langle x_k, (w^t - w^{t-1})\rangle\right\}. \quad (14)$$

By using Cauchy inequality with $\tau$, we have

$$\left\langle \sum_{k\in K}\delta_k(1-ap_k)(\alpha_k^{t+1} - \alpha_k^t)x_k, w^{t+1} - w^t \right\rangle \leq \frac{\|w^{t+1} - w^t\|_2^2}{4\tau} + \tau\sum_{k\in K}\|(\delta_k(1-ap_k))(\alpha_k^{t+1} - \alpha_k^t)x_k\|_2^2, \quad (15)$$

$$\left\langle \sum_{k\in K}\delta_k(\alpha_k^{t+1} - \alpha_k^t)x_k, w^t - w^{t-1} \right\rangle \leq \frac{\|w^t - w^{t-1}\|_2^2}{4\tau} + \tau\sum_{k\in K}\|\delta_k(\alpha_k^{t+1} - \alpha_k^t)x_k\|_2^2. \quad (16)$$

From (13), (14), (15), and (16), we have

$$\frac{\|w^t - w^*\|_2^2}{2\tau} + \sum_{i\in[n]}\left\{\left(\frac{1}{2\sigma_i} + \frac{(1-ap_i)\gamma}{p_i n}\right)\frac{(\alpha_i^t - \alpha_i^*)^2}{a}\right\} + \theta\frac{\|w^t - w^{t-1}\|_2^2}{4\tau} - \theta\frac{(\alpha^t - \alpha^*)^\top X(w^t - w^{t-1})}{n}$$
$$\geq \left(\frac{1}{2\tau} + \lambda\right)\mathbb{E}\left[\|w^{t+1} - w^*\|_2^2\right] + \sum_{i\in[n]}\left\{\left(\frac{1}{2\sigma_i} + \frac{\gamma}{p_i n}\right)\frac{\mathbb{E}\left[(\alpha_i^{t+1} - \alpha_i^*)^2\right]}{a}\right\} + \frac{\|w^{t+1} - w^t\|_2^2}{4\tau}$$
$$+ \mathbb{E}\left[\sum_{k\in K}\left(\frac{1}{2a\sigma_k} - \frac{\tau\|x_k\|_2^2((1-ap_k)^2 + \theta)}{(ap_k n)^2}\right)(\alpha_k^{t+1} - \alpha_k^t)^2\right] - \frac{(\alpha^{t+1} - \alpha^*)^\top X(w^{t+1} - w^t)}{n}.$$

From the definition of $\theta$, if the parameters $\tau, \sigma_i, \theta$ and $p$ satisfy (6), then we have $\mathbb{E}[\Delta_{t+1}] \leq \Delta_t$. ∎

## A.2 Proof of Theorem 4

**Lemma 14.** *Suppose that the assumptions of Lemma 3 hold. If $\tau, \sigma_i, \theta$, and $p$ satisfies the following inequalities:*

$$\tau\sigma_i \leq \frac{a(p_i n)^2}{4\|x_i\|_2}, \quad \frac{\sum_{k\in[n]}\|x_k\|_2^2}{n^2/\tau} \leq \frac{1}{4a\sigma_i} \quad \forall i \in [n], \tag{17}$$

*then we have*

$$\left(\frac{1}{2\tau}+\lambda\right)\mathbb{E}\left[\|w^t - w^*\|_2^2\right] + \sum_{i\in[n]}\left\{\left(\frac{1}{4\sigma_i}+\frac{\gamma}{p_i n}\right)\frac{\mathbb{E}\left[(\alpha_i^t - \alpha_i^*)^2\right]}{a}\right\} \leq \theta^t \Delta_0.$$

*Proof.* By using Cauchy inequality with $\tau$ and $\frac{\sum_{k\in[n]}\|x_k\|_2^2}{n^2/\tau} \leq \frac{1}{4a\sigma_i}$, we have

$$\frac{(\alpha^t - \alpha^*)^\top X(w^t - w^{t-1})}{n} \leq \frac{\|w^t - w^{t-1}\|_2^2}{4\tau} + \frac{\|\alpha^t - \alpha^*\|_2^2 \|X\|_F^2}{n^2/\tau}$$

$$\leq \frac{\|w^t - w^{t-1}\|_2^2}{4\tau} + \frac{\|\alpha^t - \alpha^*\|_2^2}{4a\sigma_i}. \tag{18}$$

On the other hand, we have

$$(6) \Leftrightarrow \frac{a}{2} \geq \frac{\tau\sigma_i \|x_i\|_2^2 ((1-ap_i)^2 + \theta)}{(p_i n)^2} \Leftrightarrow \tau\sigma_i \leq \frac{a(p_i n)^2}{2\|x_i\|_2^2 ((1-ap_i)^2 + \theta)}. \tag{19}$$

Form the definition $\theta$ and (19), we have

$$\tau\sigma_i \leq \frac{a(p_i n)^2}{4\|x_i\|_2^2} \leq \frac{a(p_i n)^2}{2\|x_i\|_2^2 ((1-ap_i)^2 + \theta)}. \tag{20}$$

From (20), we have

$$\frac{n^2}{\tau} \geq \frac{4\sigma_i \|x_i\|_2^2}{ap_i^2} \implies (6). \tag{21}$$

By using Lemma 3, (18), and (21), we have

$$\left(\frac{1}{2\tau}+\lambda\right)\mathbb{E}\left[\|w^t - w^*\|_2^2\right] + \sum_{i\in[n]}\left\{\left(\frac{1}{4\sigma_i}+\frac{\gamma}{p_i n}\right)\frac{\mathbb{E}\left[(\alpha_i^t - \alpha_i^*)^2\right]}{a}\right\} \leq \theta^t \Delta_0.$$

∎

**Theorem 4.** *Lemma 3 hold. For any proper distribution $p$ with $p_i \in (0, 1/a), \forall i \in [n]$, if we set the parameters $\tau$ and $\sigma_i$ as*

$$\tau = \frac{aR}{2}\sqrt{\frac{\gamma}{\lambda}}, \quad \sigma_i = \frac{np_i}{2\|x_i\|_2}\sqrt{\frac{\lambda}{\gamma}}, \quad \text{where } \underline{R} := \min_{i\in[n]}\frac{p_i}{\|x_i\|_2},$$

*then Algorithm 1 guarantees*

$$T \geq \max_{i\in[n]}\left(\frac{1}{ap_i} + \frac{\|x_i\|_2}{p_i a\sqrt{\gamma\lambda}}\right)\log\left(\frac{\Delta_0/((2\tau)^{-1}+\lambda)}{\varepsilon}\right) \implies \mathbb{E}\left[\|w^T - w^*\|_2^2\right] \leq \varepsilon.$$

*Proof.* From the limit to $p$ and the definition of $\tau, \sigma_i$, and $\underline{R}$ we have

$$\tau\sigma_i = \frac{anp_i\underline{R}}{4\|x_i\|_2} \leq \frac{an(p_i)^2}{4\|x_i\|_2^2} \leq \frac{a(p_in)^2}{4\|x_i\|_2^2}, \tag{22}$$

$$\frac{\sum_{k\in[n]}\|x_k\|_2^2}{n^2/\tau} \leq \frac{n\max_{k\in[n]}\|x_k\|_2^2}{4n\sigma_i\|x_i\|_2^2}ap_i^2 \leq \frac{1}{4a\sigma_i}. \tag{23}$$

By using (22), (23), and Lemma 14, we have

$$\left(\frac{1}{2\tau}+\lambda\right)\mathbb{E}\left[\|w^t-w^*\|_2^2\right] + \sum_{i\in[n]}\left\{\left(\frac{1}{4\sigma_i}+\frac{\gamma}{p_in}\right)\frac{\mathbb{E}\left[(\alpha_i^t-\alpha_i^*)^2\right]}{a}\right\} \leq \theta^t\Delta_0.$$

On the other hand, we have

$$1+\frac{1}{2\tau\lambda} = 1+\max_{i\in[n]}\frac{\|x_i\|_2}{p_ia\sqrt{\lambda\gamma}},$$

$$\frac{1}{ap_i}+\frac{n}{2a\gamma\sigma_i} = \frac{1}{ap_i}+\frac{\|x_i\|_2}{p_ia\sqrt{\lambda\gamma}}.$$

Therefore, the iteration $T$ which archives $\theta^T\Delta_0/((2\tau)^{-1}+\lambda) \leq \varepsilon$ is as follows:

$$T \geq \frac{\log((\Delta_0/((2\tau)^{-1}+\lambda))/\varepsilon)}{-\log(\theta)} \geq \max_{i\in[n]}\left(\frac{1}{ap_i}+\frac{\|x_i\|_2}{p_ia\sqrt{\gamma\lambda}}\right)\log\left(\frac{\Delta_0/((2\tau)^{-1}+\lambda)}{\varepsilon}\right).$$

∎

## A.3 Proof of Theorem 5

**Theorem 5.** *Suppose that the assumptions of Lemma 3 hold and $a \leq \sqrt{n}$. For any proper distribution $p$ with $p_i \in (0, 1/(a\sqrt{n})], \forall i \in [n]$, if we set the parameters $\tau$ and $\sigma_i$ as*

$$\tau = \frac{a\underline{R}}{2}\sqrt{\frac{n\gamma}{\lambda}}, \quad \sigma_i = \frac{np_i}{2\|x_i\|_2}\sqrt{\frac{n\lambda}{\gamma}}, \quad \text{where } \underline{R} := \min_{i\in[n]}\frac{p_i}{\|x_i\|_2},$$

*then Algorithm 1 guarantees for*

$$T \geq \max_{i\in[n]}\left(\frac{1}{ap_i}+\frac{\|x_i\|_2}{p_ia\sqrt{n\gamma\lambda}}\right)\log\left(\frac{\Delta_0/((2\tau)^{-1}+\lambda)}{\varepsilon}\right) \implies \mathbb{E}\left[\|w^T-w^*\|_2^2\right] \leq \varepsilon.$$

*Proof.* From the limit to $p$ and the definition of $\tau, \sigma_i$, and $\underline{R}$ we have

$$\tau\sigma_i = \frac{an^2p_i\underline{R}}{4\|x_i\|_2} \leq \frac{a(p_in)^2}{4\|x_i\|_2^2}, \tag{24}$$

$$\frac{\sum_{k\in[n]}\|x_k\|_2^2}{n^2/\tau} \leq \frac{n\max_{k\in[n]}\|x_k\|_2^2}{4\sigma_i\|x_i\|_2^2}ap_i^2 \leq \frac{1}{4a\sigma_i}. \tag{25}$$

By using (24), (25), and Lemma 14, we have

$$\left(\frac{1}{2\tau}+\lambda\right)\mathbb{E}\left[\|w^t-w^*\|_2^2\right] + \sum_{i\in[n]}\left\{\left(\frac{1}{4\sigma_i}+\frac{\gamma}{p_in}\right)\frac{\mathbb{E}\left[(\alpha_i^t-\alpha_i^*)^2\right]}{a}\right\} \leq \theta^t\Delta_0.$$

On the other hand, we have

$$1 + \frac{1}{2\tau\lambda} = 1 + \max_{i\in[n]} \frac{\|x_i\|_2}{p_i a\sqrt{n\lambda\gamma}},$$

$$\frac{1}{ap_i} + \frac{n}{2a\gamma\sigma_i} = \frac{1}{ap_i} + \frac{\|x_i\|_2}{p_i a\sqrt{n\lambda\gamma}}.$$

Therefore, the iteration $T$ which archives $\theta^T \Delta_0 / ((2\tau)^{-1} + \lambda) \leq \varepsilon$ is as follows:

$$T \geq \frac{\log((\Delta_0/((2\tau)^{-1}+\lambda))/\varepsilon)}{-\log(\theta)} \geq \max_{i\in[n]} \left(\frac{1}{ap_i} + \frac{\|x_i\|_2}{p_i a\sqrt{n\gamma\lambda}}\right) \log\left(\frac{\Delta_0/((2\tau)^{-1}+\lambda)}{\varepsilon}\right).$$

If $p$ is uniform, we have

$$\max_{i\in[n]} \left(\frac{1}{ap_i} + \frac{\|x_i\|_2}{p_i a\sqrt{n\gamma\lambda}}\right) = \frac{n}{a} + \frac{\max_{i\in[n]}\|x_i\|_2 \sqrt{n}}{a\sqrt{\lambda\gamma}}.$$

Therefore, Theorem 5 improves the dominant term of the complexity in Corollary 1 in (Zhang and Xiao, 2015) for $a \leq \sqrt{n}$.

We consider the setting of $p$ proposed in §4.1:

$$\forall i \in [n], \ p_i = \begin{cases} \frac{\rho_i \|x_i\|_2}{\sum_{k\in[n]} \rho_k \|x_k\|_2} & (\overline{p} \leq 1/(a\sqrt{n})) \\ \frac{1-\zeta}{n} + \zeta \frac{\rho_i \|x_i\|_2}{\sum_{k\in[n]} \rho_k \|x_k\|_2} & (\text{otherwise}) \end{cases}, \text{ where } \overline{p} := \max_{i\in[n]} \frac{\rho_i \|x_i\|_2}{\sum_{k\in[n]} \rho_k \|x_k\|_2}, \ \zeta := \frac{1/(a\sqrt{n}) - 1/n}{\overline{p} - 1/n}.$$

If $\overline{p} \leq 1/a$, then we have $p \in (0, 1/(a\sqrt{n})]^n$ by the definition of $\overline{p}$. In contrast, if $\overline{p} > 1/(a\sqrt{n})$, then we have also $p \in (0, 1/(a\sqrt{n})]^n$ because $p_i \leq \frac{1-\zeta}{n} + \zeta\overline{p} = 1/(a\sqrt{n})$ for all $i \in [n]$. ∎

### A.4 Proof of Theorem 15

**Theorem 15.** *Suppose that the assumptions of Lemma 3 hold and $a \geq \sqrt{n}$ in Algorithm 1. For the uniform probability vector $p$, if we set the parameters $\tau$ and $\sigma_i$ as $\tau = \frac{1}{2R}\sqrt{\frac{\gamma}{\lambda}}$, $\sigma_i = \frac{n}{2a\|x_i\|_2}\sqrt{\frac{\lambda}{\gamma}}$, where $R := \max_{i\in[n]} \|x_i\|_2$, then Algorithm 1 guarantees $\mathbb{E}\left[\|w^T - w^*\|_2^2\right] \leq \varepsilon$ for $T \geq \max_{i\in[n]} \left(\frac{n}{a} + \frac{R}{\sqrt{\gamma\lambda}}\right) \log\left(\frac{\Delta_0/((2\tau)^{-1}+\lambda)}{\varepsilon}\right)$ iterations.*

*Proof.* From the limit to $p$, limit to $a$, the definition of $\tau, \sigma_i$, and $\underline{R}$ we have

$$\tau\sigma_i = \frac{n}{4aR\|x_i\|_2} \leq \frac{a(p_i n)^2}{4\|x_i\|_2^2}, \tag{26}$$

$$\frac{\sum_{k\in[n]} \|x_k\|_2^2}{n^2/\tau} \leq \frac{nR^2}{4an\sigma_i R\|x_i\|_2} \leq \frac{1}{4a\sigma_i}. \tag{27}$$

By using (26), (27), and Lemma 14, we have

$$\left(\frac{1}{2\tau} + \lambda\right) \mathbb{E}\left[\|w^t - w^*\|_2^2\right] + \sum_{i\in[n]} \left\{\left(\frac{1}{4\sigma_i} + \frac{\gamma}{p_i n}\right) \frac{\mathbb{E}\left[(\alpha_i^t - \alpha_i^*)^2\right]}{a}\right\} \leq \theta^t \Delta_0.$$

On the other hand, we have

$$1 + \frac{1}{2\tau\lambda} = 1 + \frac{R}{\sqrt{\lambda\gamma}},$$

$$\frac{1}{ap_i} + \frac{n}{2a\gamma\sigma_i} = \frac{n}{a} + \frac{\|x_i\|_2}{\sqrt{\lambda\gamma}}$$

Therefore, the iteration $T$ which archives $\theta^T \Delta_0/((2\tau)^{-1} + \lambda) \leq \varepsilon$ is as follows:

$$T \geq \frac{\log((\Delta_0/((2\tau)^{-1} + \lambda))/\varepsilon)}{-\log(\theta)} \geq \left(\frac{n}{a} + \frac{R}{\sqrt{\lambda\gamma}}\right) \log\left(\frac{\Delta_0/((2\tau)^{-1} + \lambda)}{\varepsilon}\right).$$

∎

## A.5 Proof of Proposition 11

**Proposition 11.** *In at the iteration $t$ of algorithm 1, if $f_i$ is smoothed hinge loss and $\gamma - p_i n/\sigma_i \neq 0$ then $|-\alpha_i^t - \nabla f_i(x_i^\top \bar{w}^t)| = 0$ is a necessary and sufficient condition for $\alpha_i^{t+1} = \alpha_i^t$.*

*Proof.* We first prove $\alpha_i^{t+1} = \alpha_i^t \implies |-\alpha_i^t - \nabla f_i(x_i^\top \bar{w}^t)| = 0$. Since

$$\alpha_i^{t+1} = \underset{\beta \in \mathbb{R}}{\operatorname{argmax}} \left\{-\beta x_i^\top \bar{w}^t - f_i^*(-\beta) - \frac{p_i n}{2\sigma_i}(\beta - \alpha_i^t)^2\right\}$$

, we have

$$x_i^\top \bar{w}^t + \frac{p_i n}{\sigma_i}(\alpha_i^{t+1} - \alpha_i^t) \in \partial f_i^*(-\alpha_i^{t+1}). \tag{28}$$

From (28) and the assumption $\alpha_i^{t+1} = \alpha_i^t$ and the property of subgradient, we have $x_i^\top \bar{w}^t + \in \partial f_i^*(-\alpha_i^{t+1}) \Leftrightarrow -\alpha_i^t \in \partial f_i(x_i^\top \bar{w}^t)$. Therefore, if assumption holds, then $\alpha_i^{t+1} = \alpha_i^t \implies -\alpha_i^t - \nabla f_i(x_i^\top \bar{w}^t) = 0$, because $f_i$ is differentiable. Next, we prove $-\alpha_i^t - \nabla f_i(x_i^\top \bar{w}^t) = 0 \implies \alpha_i^{t+1} = \alpha_i^t$. Since $-\alpha_i^t = \nabla f_i(x_i^\top \bar{w}^t)$, and from the property of subgradient, we have $x_i^\top \bar{w}^t = \partial f_i^*(-\alpha_i^t)$. From (28), we have $0 \in \partial f_i^*(-\alpha_i^{t+1}) - \partial f_i^*(-\alpha_i^t) - \frac{p_i n}{\sigma_i}(\alpha_i^{t+1} - \alpha_i^t)$. From the differential of the conjugate of smoothed hinge loss, we have $(\gamma - p_i n/\sigma_i)(\alpha_i^{t+1} - \alpha_i^t) = 0$ Therefore, if the assumptions hold, then we have $\alpha_i^{t+1} = \alpha_i^t$. ∎

## A.6 Proof of Proposition 12

**Proposition 12.** *In at the iteration $t$ of algorithm 1, if $g$ is elastic net $\lambda + 1/\tau \neq 0$ and $sign(w_j^{t+1}) = sign(w_j^t)$ then $|w_j^t - \nabla g_j^*\left(\frac{1}{\lambda n} X_{:j}^\top \bar{\alpha}^{t+1}\right)| = 0$ is a necessary and sufficient condition for $w_j^{t+1} = w_j^t$.*

*Proof.* We first prove $w_j^{t+1} = w_j^t \implies |w_j^t - \nabla g_j^*\left(\frac{1}{\lambda n} X_{:j}^\top \bar{\alpha}^{t+1}\right)| = 0$. Since

$$w_j^{t+1} = \underset{v \in \mathbb{R}}{\operatorname{argmin}} \left\{\lambda g_j(v) - \frac{v}{n}\left\langle X_{:j}, \bar{\alpha}^{t+1}\right\rangle + \frac{1}{2\tau}(v - w_j^t)^2\right\}$$

, we have

$$\frac{1}{\lambda n} X_{:j}^\top \bar{\alpha}^{t+1} - \frac{1}{\lambda \tau}(w_j^{t+1} - w_j^t) \in \partial g_j(w_j^{t+1}). \tag{29}$$

From (29) and the assumption $w_j^{t+1} = w_j^t$ and the property of subgradient, we have $w_j^{t+1} \in \partial g_j^*(\frac{1}{\lambda n} X_{:j}^\top \bar{\alpha}^{t+1})$. Therefore, if the assumptions hold, then $w_j^{t+1} = w_j^t \implies w_j^{t+1} - \nabla g_j^*(\frac{1}{\lambda n} X_{:j}^\top \bar{\alpha}^{t+1}) = 0$, because $g_j^*$ is differentiable. Next, we prove $w_j^{t+1} - \nabla g_j^*(\frac{1}{\lambda n} X_{:j}^\top \bar{\alpha}^{t+1}) = 0 \implies w_j^{t+1} = w_j^t$. Since $w_j^{t+1} = \nabla g_j^*(\frac{1}{\lambda n} X_{:j}^\top \bar{\alpha}^{t+1})$, and from the property of subgradient, we have $\frac{1}{\lambda n} X_{:j}^\top \bar{\alpha}^{t+1} \in \partial g_j(w_j^t)$. From (29), we have $0 \in \partial g_j(w_j^{t+1}) - \partial g_j(w_j^t) + \frac{1}{\lambda \tau}(w_j^{t+1} - w_j^t)$. From the differential of elastic net penalty and the assumptions, we have $(\lambda + 1/\tau)(w_j^{t+1} - w_j^t) = 0$ Therefore, if the assumptions hold, then we have $w_j^{t+1} = w_j^t$. ∎

## B SPDC v.s. Quartz in mini-batching with the uniform sampling

In this Appendix, we compare our results with SPDC (Zhang and Xiao, 2015) and Quartz (Qu et al., 2015) using the uniform sampling in mini-batch setting. Table 5 shows the dominant factor of the iteration complexity of SPDC and Quartz when mini-batch size $a > 1$ and $\|x_i\|_2 = 1$ for all $i \in [n]$. Table 6 shows the dominant factor of the iteration complexity when $\lambda\gamma n = \Theta(1/\sqrt{n})$ or $\lambda\gamma n = \Theta(1)$ or $\lambda\gamma n = \Theta(\sqrt{n})$. We can see that the iteration complexity of (Zhang and Xiao, 2015) is decreased by only $1/\sqrt{a}$ times when $\lambda\gamma n = \Theta(1/\sqrt{n})$ and $\lambda\gamma n = \Theta(1)$, although the iteration complexities of Theorem 5 and 15 are decreased by $1/a$ times. We demonstrate the advantage of our results through numerical experiments. The experimental setups are the same as those in §4.4. Table 7 shows the datasets used in the experiments. Figure 5 illustrates the results when $\lambda\gamma n = 1/\sqrt{n}$, $\lambda\gamma n = 1$, and $\lambda\gamma n = \sqrt{n}$. We can see that the our results are more effective than the previous SPDC (Zhang and Xiao, 2015). Quartz (Qu et al., 2015) can converge on the artificial data whose $\tilde{r}$ is extremely small, although Quartz converge slowly on `rcv1-train` and `real-sim`.

| Algorithm | Dominant factor of iteration complexity |
|---|---|
| Quartz (Qu et al., 2015) | $\frac{n}{a} + \left(1 + \frac{(\tilde{r}-1)(a-1)}{n-1}\right)\frac{1}{\lambda\gamma n}$ |
| SPDC (Zhang and Xiao, 2015) | $\frac{n}{a} + \sqrt{\frac{n}{\lambda\gamma a}}$ |
| SPDC (Theorem 5 and 15 in this paper) | $\frac{n}{a} + \max\left\{\frac{1}{a}\sqrt{\frac{n}{\lambda\gamma}}, \sqrt{\frac{1}{\lambda\gamma}}\right\}$ |

Table 5: The dominant factor of the iteration complexity of SPDC and Quartz when mini-batch size $a > 1$ and $\|x_i\|_2 = 1$ for all $i \in [n]$. We define $\tilde{r}$ as $\tilde{r} := 1 + (B_{\tilde{i}}/\max_{i\in[n]}\|x_i\|_2^2 - 1)(n-1)/(a-1)$, where $B_i := \left\{\sum_{j=1}^d (1 + (r_j - 1)(a-1)/(n-1))X_{ij}^2\right\}$ for all $i \in [n]$, $r_j := |i \in [n]|X_{ij} \neq 0|$ for all $j \in [d]$, and $\tilde{i} := \operatorname*{argmax}_{i\in[n]} B_i$.

| Algorithm | | Dominant factor of iteration complexity | | |
|---|---|---|---|---|
| | | $\lambda\gamma n = \Theta(1/\sqrt{n})$ | $\lambda\gamma n = \Theta(1)$ | $\lambda\gamma n = \Theta(\sqrt{n})$ |
| Quartz (Qu et al., 2015) | | $O(\frac{n^{3/2}}{a} + \tilde{r}\sqrt{n})$ | $O(\frac{n}{a} + \tilde{r})$ | $O(\frac{n}{a} + \frac{\tilde{r}}{\sqrt{n}})$ |
| SPDC (Zhang and Xiao, 2015) | | $O(\frac{n^{5/4}}{\sqrt{a}})$ | $O(\frac{n}{\sqrt{a}})$ | $O(\frac{n^{3/4}}{\sqrt{a}} + \frac{n}{a})$ |
| SPDC (This paper) | $a \leq \sqrt{n}$ | $O(n^{5/4}/a)$ | $O(n/a)$ | $O(n/a)$ |
| | $a \geq \sqrt{n}$ | $O(n/a + n^{3/4})$ | $O(n/a + n^{1/2})$ | $O(n/a + n^{1/4})$ |

Table 6: Comparison of the dominant factor of the iteration complexity of SPDC and Quartz when $\lambda\gamma n = \Theta(1/\sqrt{n})$ or $\lambda\gamma n = \Theta(1)$ or $\lambda\gamma n = \Theta(\sqrt{n})$. We note $\tilde{r} = n$ when #(nnz)/nd = 1.

## C AdaSPDC with data-driven sampling

In this Appendix, we derive the iteration complexity of AdaSPDC with data-driven sampling when $a = 1$. The following corollaries can derived by using Theorem 4 and 5.

**Corollary 16.** *Suppose that the assumptions of Lemma 3 hold and $a = 1$ in Algorithm 1. For the probability vector $p$ with $p_i = (\|x_i\|_2 + \sqrt{\gamma\lambda n})/(\sum_{k\in[n]}\|x_k\|_2 + \sqrt{\gamma\lambda n}), \forall i \in [n]$, if we set the parameters $\tau$ and $\sigma_i$ as*

| Dataset name | $\tilde{r}$ | $\min_{j \in [d]} r_j$ | $\max_{j \in [d]} r_j$ | mean($r_j$) | vari($r_j$) | #(nnz)/$nd$ |
|---|---|---|---|---|---|---|
| Artificial data | 1.99 ($a=4$) <br> 1.99 ($a=16$) | 0 | 2 | 0.102 | 0.099 | 0.000105 |
| rcv1-train | 2130 ($a=10$) <br> 2130 ($a=100$) | 0 | 8551 | 31.73 | 39735 | 0.001567 |
| real-sim | 19982 ($a=10$) <br> 19982 ($a=100$) | 11 | 41607 | 177 | 468075 | 0.002451 |

Table 7: Benchmark datasets used in the experiments. We can bound $\tilde{r}$ as $1 \leq \tilde{r} \leq \max_{j \in [d]} r_j \leq n$.

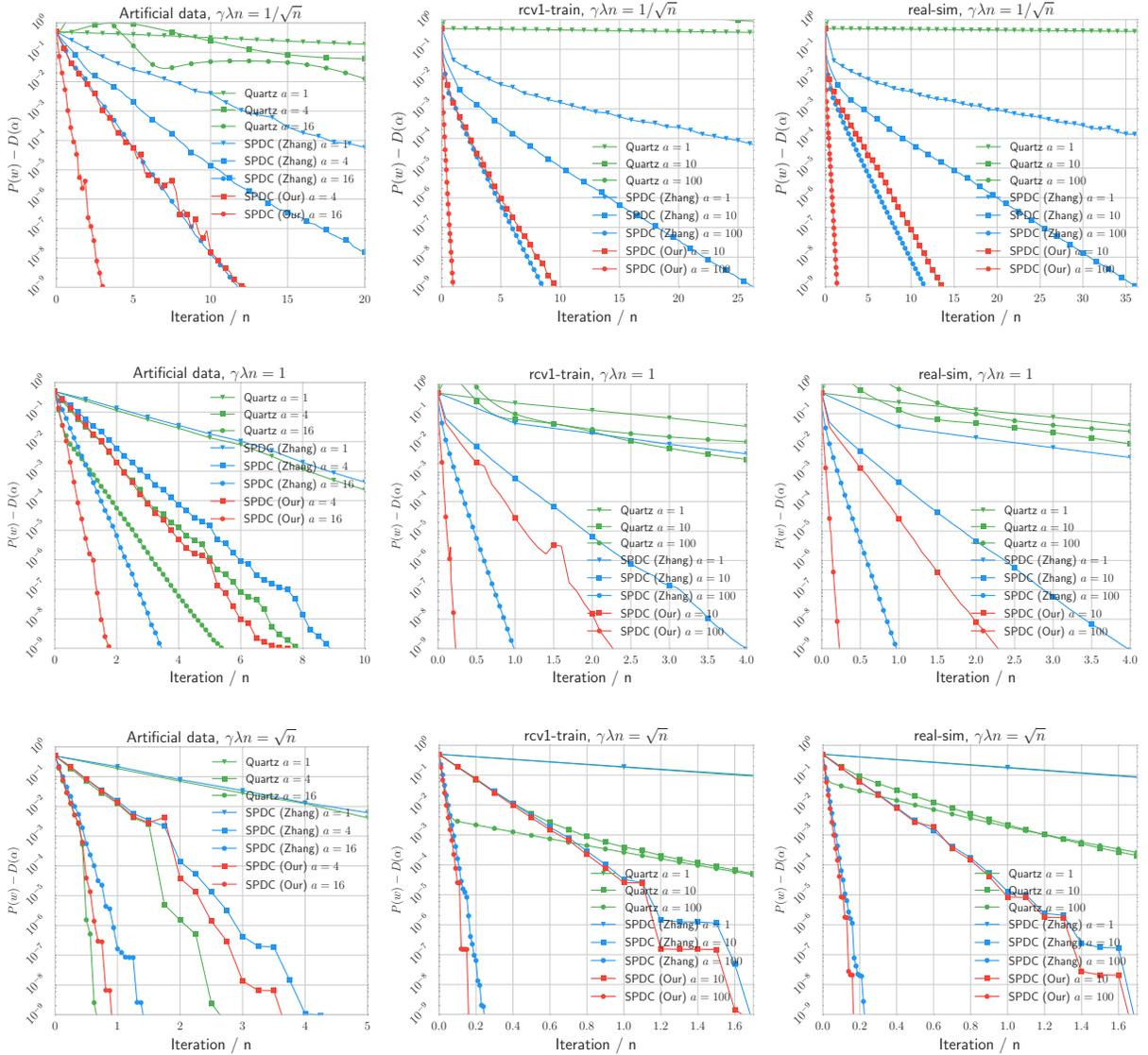

Figure 5: The illustration of the convergence speed of each method on each dataset described in Table 7.

$\tau = \frac{R}{2}\sqrt{\frac{\gamma}{\lambda}}, \sigma_i = \frac{np_i}{2\|x_i\|_2}\sqrt{\frac{\lambda}{\gamma}}$, *where* $\underline{R} := \min_{i\in[n]} \frac{p_i}{\|x_i\|_2}$, *then Algorithm 1 guarantees* $\mathbb{E}\left[\|w^T - w^*\|_2^2\right] \leq \varepsilon$ *for* $T \geq \left(n + \frac{\sum_{k\in[n]}\|x_k\|_2}{\sqrt{\gamma\lambda}}\right)\log\left(\frac{\Delta_0/((2\tau)^{-1}+\lambda)}{\varepsilon}\right)$ *iterations.*

**Corollary 17.** *Suppose that the assumptions of Lemma 3 and* $\max_{i\in[n]}\|x_i\|_2 \leq \frac{1}{\sqrt{n}}(\sum_{k\in[n]}\|x_k\|_2) + \sqrt{\lambda\gamma}(n-\sqrt{n})$ *hold and* $a = 1$ *in Algorithm 1. For the probability vector $p$ with* $p_i = (\|x_i\|_2 + \sqrt{\gamma\lambda n})/(\sum_{k\in[n]}\|x_k\|_2 + \sqrt{\gamma\lambda n}), \forall i \in [n]$, *if we set the parameters $\tau$ and $\sigma_i$ as* $\tau = \frac{R}{2}\sqrt{\frac{n\gamma}{\lambda}}, \sigma_i = \frac{np_i}{2\|x_i\|_2}\sqrt{\frac{n\lambda}{\gamma}}$, *where* $\underline{R} := \min_{i\in[n]} \frac{p_i}{\|x_i\|_2}$, *then Algorithm 1 guarantees* $\mathbb{E}\left[\|w^T - w^*\|_2^2\right] \leq \varepsilon$ *for* $T \geq \left(n + \frac{\sum_{k\in[n]}\|x_k\|_2}{\sqrt{n\gamma\lambda}}\right)\log\left(\frac{\Delta_0/((2\tau)^{-1}+\lambda)}{\varepsilon}\right)$ *iterations.*

| SPDC with data-driven sampling | $p_i$ | Dominant factor of iteration complexity | Limit to $\max_{i\in[n]}\|x_i\|_2$ |
|---|---|---|---|
| Theorem 2 in (Zhang and Xiao, 2015) | $\frac{1}{2n} + \frac{\|x_i\|_2}{2\bar{R}}$ | $2n + 2\frac{\bar{R}}{\sqrt{\lambda\gamma n}}$ | - |
| Corollary 16 in this paper | $\frac{\|x_i\|_2 + \sqrt{\lambda\gamma}}{\bar{R} + n\sqrt{\lambda\gamma}}$ | $n + \frac{\bar{R}}{\sqrt{\lambda\gamma}}$ | - |
| Corollary 17 in this paper | $\frac{\|x_i\|_2 + \sqrt{n\lambda\gamma}}{\bar{R} + n\sqrt{\lambda\gamma n}}$ | $n + \frac{\bar{R}}{\sqrt{\lambda\gamma n}}$ | $\leq \frac{\bar{R}}{\sqrt{n}} + \sqrt{\lambda\gamma}(n - \sqrt{n})$ |

Table 8: List of the dominant factor of the iteration complexity of SPDC with data-driven sampling. We define $\bar{R}$ as $\bar{R} := \sum_{i\in[n]}\|x_i\|_2$.

Table 8 shows the dominant factor of the iteration complexity of Weighted SPDC (Zhang and Xiao, 2015) and our methods. We demonstrate the advantage of our results through numerical experiments. The experimental setups are the same as those in §4.4. Figure 6 shows the results. We can see that the our data-driven sampling is more effective than Weighted SPDC (Zhang and Xiao, 2015).

## D vanilla SPDC with arbitrary sampling

In this appendix, we derive the iteration complexity of vanilla SPDC with arbitrary sampling. We omit the proofs because the following corollaries are almost same as Theorem 4 and 5.

**Corollary 18.** *Suppose that the assumptions of Lemma 3 hold. For any proper distribution $p$ with $p_i \in (0, 1/a), \forall i \in [n]$, if we set the parameters $\tau$ and $\sigma_i$ as* $\tau = \frac{aR}{2}\sqrt{\frac{\gamma}{\lambda}}, \sigma = \frac{nR}{2}\sqrt{\frac{\lambda}{\gamma}}$, *where* $\underline{R} := \min_{i\in[n]} \frac{p_i}{\|x_i\|_2}$, *then Algorithm 4 guarantees* $\mathbb{E}\left[\|w^T - w^*\|_2^2\right] \leq \varepsilon$ *for* $T \geq \max_{i\in[n]}\left(\frac{1}{ap_i} + \frac{1}{\underline{R}a\sqrt{\gamma\lambda}}\right)\log\left(\frac{\Delta_0/((2\tau)^{-1}+\lambda)}{\varepsilon}\right)$ *iterations.*

**Corollary 19.** *Suppose that the assumptions of Lemma 3 hold and $a \leq \sqrt{n}$. For any proper distribution $p$ with $p_i \in (0, 1/(a\sqrt{n})], \forall i \in [n]$, if we set the parameters $\tau$ and $\sigma_i$ as* $\tau = \frac{aR}{2}\sqrt{\frac{n\gamma}{\lambda}}, \sigma = \frac{nR}{2}\sqrt{\frac{n\lambda}{\gamma}}$, *where* $\underline{R} := \min_{i\in[n]} \frac{p_i}{\|x_i\|_2}$, *then Algorithm 4 guarantees* $\mathbb{E}\left[\|w^T - w^*\|_2^2\right] \leq \varepsilon$ *for* $T \geq \max_{i\in[n]}\left(\frac{1}{ap_i} + \frac{1}{\underline{R}a\sqrt{n\gamma\lambda}}\right)\log\left(\frac{\Delta_0/((2\tau)^{-1}+\lambda)}{\varepsilon}\right)$ *iterations.*

## E Doubly SPDC with arbitrary sampling

Here, we analyze Doubly SPDC (DSPDC) was proposed by (Wei Yu et al., 2015). For using non-uniform sampling and the convenience of the proof, we made changes in DSPDC as Algorithm 5. Here, we assume the separability of the penalty function (i.e., $g(w) = \sum_{j\in[d]} g_j(w_j)$).

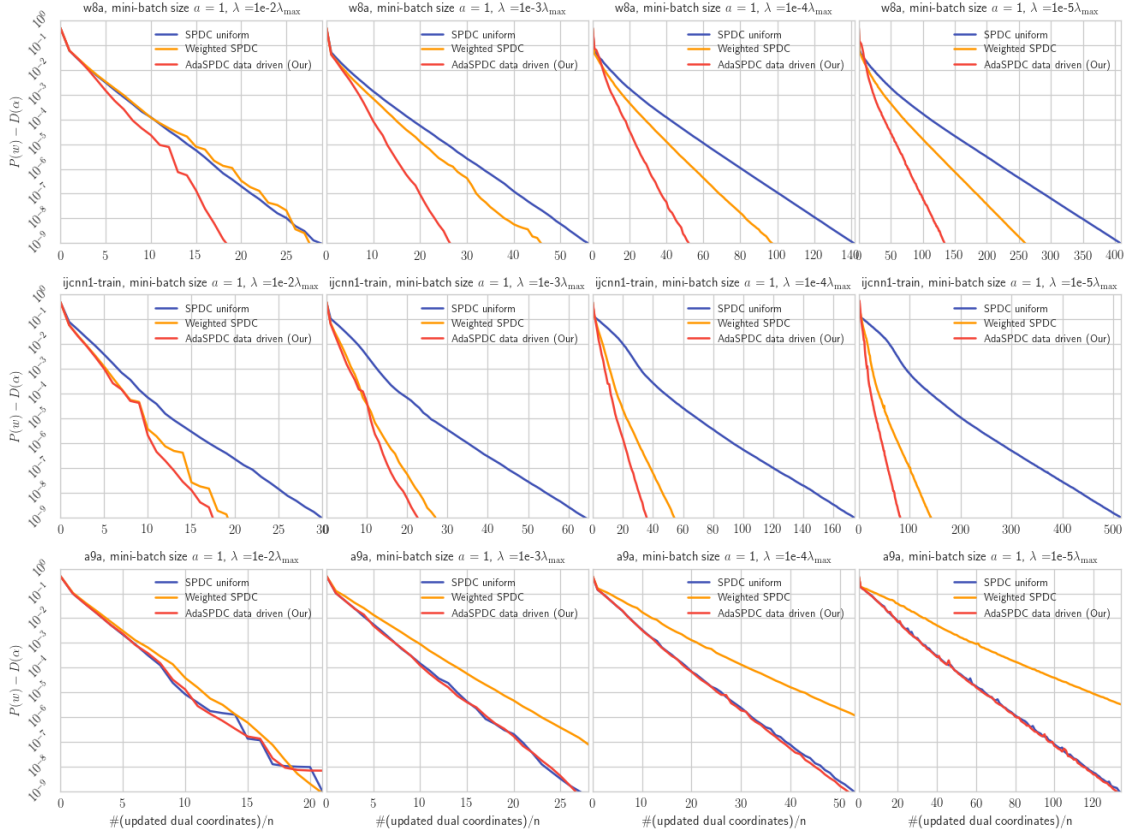

Figure 6: The illustration of the convergence speed of each method on each dataset.

---

**Algorithm 4:** (vanilla) SPDC with non-uniform sampling (Zhang and Xiao, 2015)

**Initialize**: $a$ (mini-batch size), $w^0, \alpha^0$ (initial solutions), $\bar{w}^0 \leftarrow w^0, \bar{\alpha}^0 \leftarrow \alpha^0$

**for** $t = 1, 2, \ldots$ **to** *converged* **do**

Generate random index from $p$ $a$ times with replacement (denote the set of random indices as $K$)
Update the dual and the primal coordinates:

$$\alpha_i^{t+1} = \begin{cases} \underset{\beta \in \mathbb{R}}{\text{argmax}} \left\{ -\beta \langle x_i, \bar{w}^t \rangle - f_i^*(-\beta) - \frac{p_i n}{2\sigma}(\beta - \alpha_i^t)^2 \right\} & (i \in K) \\ \alpha_i^t & (i \notin K) \end{cases}$$

$$\bar{\alpha}_i^{t+1} = \alpha_i^t + \frac{1}{a p_i n}(\alpha_i^{t+1} - \alpha_i^t) \quad (i \in [n])$$

$$w^{t+1} = \underset{w \in \mathbb{R}^d}{\text{argmin}} \left\{ \lambda g(w) - \frac{1}{n} w^\top X^\top \bar{\alpha}^{t+1} + \frac{1}{2\tau} \|w - w^t\|_2^2 \right\}$$

$$\bar{w}^{t+1} = w^{t+1} + \theta(w^{t+1} - w^t),$$

where $\theta := \max\left\{ 1 - \left(\frac{1}{2\tau\lambda}\right)^{-1}, 1 - \left(\max_{i \in [n]} \frac{1}{a p_i} + \frac{n}{2a\sigma\gamma}\right)^{-1} \right\}$, $\sigma, \tau, \theta$ are the parameters that would be uniquely determined if the probability vector $p$ is determined.

---

**Algorithm 5:** DSPDC with non-uniform sampling

**Initialize**: $a$ (dual mini-batch size), $b$ (primal mini-batch size), $w^0, \alpha^0$ (initial solutions), $\bar{w}^0 \leftarrow w^0, \bar{\alpha}^0 \leftarrow \alpha^0$
Partition the indices of primal variables into $d \bmod b$ mini-batches whose the number of elements are $b+1$ and $(\lfloor d/b \rfloor - d \bmod b)$ mini-batches whose the number of elements are $b$
**for** $t = 1, 2, \ldots$ **to** *converged* **do**
    Generate a random index from $p$ $a$ times (denote the set of random indices as $K$)
    Randomly pick up a subset $M_h$ follows the probability vector $q$
    Update primal and dual coordinates:

$$\alpha_i^{t+1} = \begin{cases} \underset{\beta \in \mathbb{R}}{\operatorname{argmax}} \left\{ -\beta \langle x_i, \bar{w}^t \rangle - f_i^*(-\beta) - \frac{p_i n}{2\sigma}(\beta - \alpha_i^t)^2 \right\} & (i \in K) \\ \alpha_i^t & (i \notin K) \end{cases}$$

$$\bar{\alpha}_i^{t+1} = \alpha_i^t + \frac{1}{ap_k n}(\alpha_i^{t+1} - \alpha_i^t) \quad (i \in [n])$$

$$w_j^{t+1} = \begin{cases} \underset{v \in \mathbb{R}}{\operatorname{argmin}} \left\{ \lambda g_j(v) - \frac{v}{n} \langle X_{:j}, \bar{\alpha}^{t+1} \rangle + \frac{q_{M_h} d}{2\tau |M_h|}(v - w_j^t)^2 \right\} & (j \in M_h) \\ w_j^t & (j \notin M_h) \end{cases}$$

$$\bar{w}^{t+1} = w^{t+1} + \theta(w^{t+1} - w^t)$$

**Theorem 20.** *Assume that $g$ is 1-strongly convex and for each $i \in [n]$, $f_i$ is $1/\gamma$-smooth and convex. Let $p \in (0, 1/a]^n$, $q \in \mathbb{R}_{++}^{\lfloor d/b \rfloor}$ be probability vectors and define $\Delta_t$ as follows:*

$$\Delta_t := \sum_{l \in \lfloor d/b \rfloor} \left\{ \left( \frac{d}{2\tau |M_l|} + \frac{\lambda}{q_{M_l}} \right) \|w_{M_l}^t - w_{M_l}^*\|_2^2 \right\} + \sum_{i \in [n]} \left\{ \left( \frac{1}{2\sigma} + \frac{\gamma}{p_i n} \right) \frac{(\alpha_i^t - \alpha_i^*)^2}{a} \right\}$$
$$+ \frac{d}{4\tau |M_h|} \|w^t - w^{t-1}\|_2^2 - \frac{1}{nq_{M_h}}(\alpha^t - \alpha^*)X(w^t - w^{t-1}).$$

*If the parameters $\tau, \sigma, \theta$ and $p, q$ in Algorithm 5 satisfy the following inequalities:*

$$\left( \frac{d}{2\tau |M_l|} + \frac{\lambda(1 - q_{M_l})}{q_{M_l}} \right) \Big/ \left( \frac{d}{2\tau |M_l|} + \frac{\lambda}{q_{M_l}} \right) \leq \bar{\theta}, \quad \left( \frac{1}{2\sigma} + \frac{(1 - ap_i)\gamma}{p_i n} \right) \Big/ \left( \frac{1}{2\sigma} + \frac{\gamma}{p_i n} \right) \leq \bar{\theta} \ (\forall i \in [n]),$$
$$\frac{\theta}{n} \Big/ \frac{1}{nq_{M_h}} \leq \bar{\theta}, \quad \frac{\theta}{4\tau} \Big/ \frac{1}{4\tau q_{M_h}} \leq \bar{\theta}, \quad \frac{1}{2a\sigma_t} - \frac{((1 - ap_k)^2/q_{M_h} + \theta)\tau \Lambda_k}{(ap_k n)^2} \geq 0 \ (\forall k \in K), \quad (30)$$

*where $\Lambda_k := \max_{|M|=b+1} \|X_{k,M}\|^2$ then for each $t \geq 0$, Algorithm 5 achieves $\mathbb{E}[\Delta_{t+1}] \leq \theta \Delta_t$.*

*Proof.* Let $\tilde{w}_j$ as follows:

$$\tilde{w}_j := \underset{v \in \mathbb{R}}{\operatorname{argmin}} \left\{ \lambda g_j(v) - \frac{v}{n} \langle X_{:j}, \bar{\alpha}^{t+1} \rangle + \frac{q_{M_h} d}{2\tau |M_h|}(v - w_j^t)^2 \right\}, \quad (j \in M_h).$$

Since the objective function of the above optimization problem is $\lambda + g_{M_h} d / \tau |M_h|$-strongly convex from 1-strongly convexity of $g_j$, , we have

$$\lambda g_j(w_j^*) - \frac{w_j^*}{n} \langle X_{:j}, \bar{\alpha}^{t+1} \rangle + \frac{q_{M_h} d}{2\tau |M_h|}(w_j^t - w_j^*)^2 \geq \tag{31}$$

$$\lambda g_j(\tilde{w}_j) - \frac{\tilde{w}_j}{n} \langle X_{:j}, \bar{\alpha}^{t+1} \rangle + \frac{q_{M_h} d}{2\tau |M_h|}(w_j^t - \tilde{w}_j)^2 + \left( \frac{\lambda}{2} + \frac{q_{M_h} d}{2\tau |M_h|} \right) (\tilde{w}_j - w_j^*)^2. \tag{32}$$

Since $g_j$ is 1-strongly convex, we have

$$\lambda g_j(\tilde{w}_j) - \frac{\tilde{w}_j}{n} \langle X_{:j}, \alpha^* \rangle \geq \lambda g_j(w_j^*) - \frac{w_j^*}{n} \langle X_{:j}, \alpha^* \rangle + \frac{\lambda}{2}(\tilde{w}_j - w_j^*)^2. \tag{33}$$

By combining (31) and (33), and summing $\forall j \in M_h$ both side, we have

$$\frac{q_{M_h}d}{2\tau|M_h|} \left\| w_{M_h}^t - w_{M_h}^* \right\|_2^2 \geq \left( \frac{g_j d}{2\tau|M_h|} + \lambda \right) \left( \frac{1}{q_{M_h}} \mathbb{E}\left[ \left\| w_{M_h}^{t+1} - w_{M_h}^* \right\|_2^2 \right] - \frac{1 - q_{M_h}}{q_{M_h}} \left\| w_{M_h}^t - w_{M_h}^* \right\|_2^2 \right)$$
$$+ \frac{d}{2\tau|M_h|} \mathbb{E}\left[ \left\| w_{M_h}^{t+1} - w_{M_h}^t \right\|_2^2 \right] - \frac{1}{n}(\bar{\alpha}^{t+1} - \alpha^*) X_{:M_j} \left( \frac{1}{q_{M_h}} \mathbb{E}\left[ w_{M_h}^{t+1} \right] - \frac{1 - q_{M_h}}{q_{M_h}} (w_{M_h}^t - w_{M_h}^*) \right),$$

because

$$\mathbb{E}\left[ \left\| w_{M_h}^{t+1} - w_{M_h}^t \right\|_2^2 \right] = q_{M_h}(\tilde{w}_{M_h} - w_{M_h}^t)^2$$

$$\mathbb{E}\left[ \left\| w_{M_h}^{t+1} - w_{M_h}^* \right\|_2^2 \right] = q_{M_h} \left\| \tilde{w}_{M_h} - w_j^* \right\|_2^2 + (1 - q_{M_h}) \left\| w_{M_h}^t - w_{M_h}^* \right\|_2^2$$

$$\mathbb{E}\left[ w_{M_h}^{t+1} \right] = q_{M_h}\tilde{w}_{M_h} - (1 - q_{M_h})w_{M_h}^t.$$

By summing $M_l = M_1, \ldots, M_{\lfloor d/b \rfloor}$,

$$\sum_{l \in \lfloor d/b \rfloor} \left\{ \left( \frac{d}{2\tau|M_l|} + \frac{\lambda(1 - q_{M_l})}{q_{M_l}} \right) \mathbb{E}\left[ \left\| w_{M_l}^t - w_{M_l}^* \right\|_2^2 \right] \right\} \geq \sum_{j \in \lfloor d/b \rfloor} \left\{ \left( \frac{d}{2\tau|M_l|} + \frac{\lambda}{q_{M_l}} \right) \mathbb{E}\left[ \left\| w_{M_l}^{t+1} - w_{M_l}^* \right\|_2^2 \right] \right\}$$
$$+ \frac{d}{2\tau|M_l|} \mathbb{E}\left[ \left\| w^{t+1} - w^t \right\|_2^2 \right] - \frac{1}{n} \mathbb{E}\left[ (\bar{\alpha}^{t+1} - \alpha^*) X \left( w^t - w^* + \frac{1}{q_{M_h}}(w^{t+1} - w^t) \right) \right] \tag{34}$$

From (10) and (34), we have

$$\sum_{l \in \lfloor d/b \rfloor} \left\{ \left( \frac{d}{2\tau|M_l|} + \frac{\lambda(1 - q_{M_l})}{q_{M_l}} \right) \mathbb{E}\left[ \left\| w_{M_l}^t - w_{M_l}^* \right\|_2^2 \right] \right\} + \sum_{i \in [n]} \left\{ \left( \frac{1}{2\sigma} + \frac{(1 - ap_i)\gamma}{p_i n} \right) \frac{(\alpha_i^t - \alpha_i^*)^2}{a} \right\}$$
$$\geq \sum_{j \in \lfloor d/b \rfloor} \left\{ \left( \frac{d}{2\tau|M_l|} + \frac{\lambda}{q_{M_l}} \right) \mathbb{E}\left[ \left\| w_{M_l}^{t+1} - w_{M_l}^* \right\|_2^2 \right] \right\} + \sum_{i \in [n]} \left\{ \left( \frac{1}{2\sigma} + \frac{\gamma}{p_i n} \right) \frac{\mathbb{E}\left[ (\alpha_i^{t+1} - \alpha_i^*)^2 \right]}{a} \right\}$$
$$+ \frac{d}{2\tau|M_l|} \mathbb{E}\left[ \left\| w^{t+1} - w^t \right\|_2^2 \right] + \frac{1}{2a\sigma} \mathbb{E}\left[ \left\| \alpha^{t+1} - \alpha^t \right\|_2^2 \right] - \frac{1}{n} \mathbb{E}\left[ (\bar{\alpha}^{t+1} - \alpha^*) X \left( \frac{1}{q_{M_h}}(w^{t+1} - w^t) - \theta(w^t - w^{t-1}) \right) \right].$$

Since

$$\frac{1}{n}((\delta - \mathbf{1}) \odot (\alpha^{t+1} - \alpha^t)) X (w^{t+1} - w^t) = \left\langle \sum_{k \in K} \frac{1 - ap_k}{ap_k n} (\alpha_k^{t+1} - \alpha_k^t) X_{kM_h}, w_{M_h}^{t+1} - w_{M_h}^t \right\rangle$$

$$\leq \tau \sum_{k \in K} \frac{\Lambda_k(1 - ap_k)^2(\alpha_k^{t+1} - \alpha_k^t)^2}{(ap_k n)^2} + \frac{\left\| w^{t+1} - w^t \right\|_2^2}{4\tau}$$

$$\frac{1}{n}(\delta \odot (\alpha^{t+1} - \alpha^t)) X (w^t - w^{t-1}) \leq \tau \sum_{k \in K} \frac{\Lambda_k(\alpha_k^{t+1} - \alpha_k^t)^2}{(ap_k n)^2} + \frac{\left\| w^t - w^{t-1} \right\|_2^2}{4\tau}$$

, we have

$$
-\frac{1}{n}(\bar{\alpha}^{t+1} - \alpha^*)X\left(\frac{1}{q_{M_h}}(w^{t+1} - w^t) - \theta(w^t - w^{t-1})\right)
$$

$$
= -\frac{1}{nq_{M_h}}(\alpha^t - \alpha^*)X(w^{t+1} - w^t) - \frac{1}{nq_{M_h}}((\delta - \mathbf{1}) \odot (\alpha^{t+1} - \alpha^t))X(w^{t+1} - w^t)
$$

$$
+ \frac{\theta}{n}(\alpha^t - \alpha^*)X(w^{t+1} - w^t) + \frac{\theta}{n}(\delta \odot (\alpha^{t+1} - \alpha^t))X(w^t - w^{t-1})
$$

$$
\geq -\frac{1}{nq_{M_h}}(\alpha^t - \alpha^*)X(w^{t+1} - w^t) + \frac{\theta}{n}(\alpha^t - \alpha^*)X(w^{t+1} - w^t)
$$

$$
- \frac{\|w^{t+1} - w^t\|_2^2}{4\tau q_{M_h}} - \theta\frac{\|w^t - w^{t-1}\|_2^2}{4\tau} - \sum_{k \in K}\left(\frac{\tau\Lambda_k((1-ap_k)^2/q_{M_h} + \theta)}{(ap_k n)^2}(\alpha_k^{t+1} - \alpha_k^t)^2\right).
$$

Therefore, we have

$$
\sum_{l \in \lfloor d/b \rfloor}\left\{\left(\frac{d}{2\tau|M_l|} + \frac{\lambda(1 - q_{M_l})}{q_{M_l}}\right)\mathbb{E}\left[\|w_{M_l}^t - w_{M_l}^*\|_2^2\right]\right\} + \sum_{i \in [n]}\left\{\left(\frac{1}{2\sigma} + \frac{(1-ap_i)\gamma}{p_i n}\right)\frac{(\alpha_i^t - \alpha_i^*)^2}{a}\right\}
$$

$$
- \frac{\theta}{n}(\alpha^t - \alpha^*)X(w^{t+1} - w^t) + \theta\frac{\|w^t - w^{t-1}\|_2^2}{4\tau}
$$

$$
\geq \sum_{j \in \lfloor d/b \rfloor}\left\{\left(\frac{d}{2\tau|M_l|} + \frac{\lambda}{q_{M_l}}\right)\mathbb{E}\left[\|w_{M_l}^{t+1} - w_{M_l}^*\|_2^2\right]\right\} + \sum_{i \in [n]}\left\{\left(\frac{1}{2\sigma} + \frac{\gamma}{p_i n}\right)\frac{\mathbb{E}\left[(\alpha_i^{t+1} - \alpha_i^*)^2\right]}{a}\right\}
$$

$$
- \frac{1}{nq_{M_h}}(\alpha^t - \alpha^*)X(w^{t+1} - w^t) + \left(\frac{d}{2\tau|M_l|} - \frac{1}{4\tau q_{M_h}}\right)\mathbb{E}\left[\|w^{t+1} - w^t\|_2^2\right]
$$

$$
+ \sum_{k \in K}\left\{\left(\frac{1}{2a\sigma} - \frac{\tau\Lambda_k((1-ap_k)^2/q_{M_h} + \theta)}{(ap_k n)^2}\right)\mathbb{E}\left[(\alpha_k^{t+1} - \alpha_k^t)^2\right]\right\}.
$$

Hence, if the inequalities (30) satisfy, then we have $\mathbb{E}\left[\Delta_{t+1}\right] \leq \Delta_t$. ∎

## F Other experiments

In this appendix, we show the rest of the experimental results in §4.4 and 5.1.

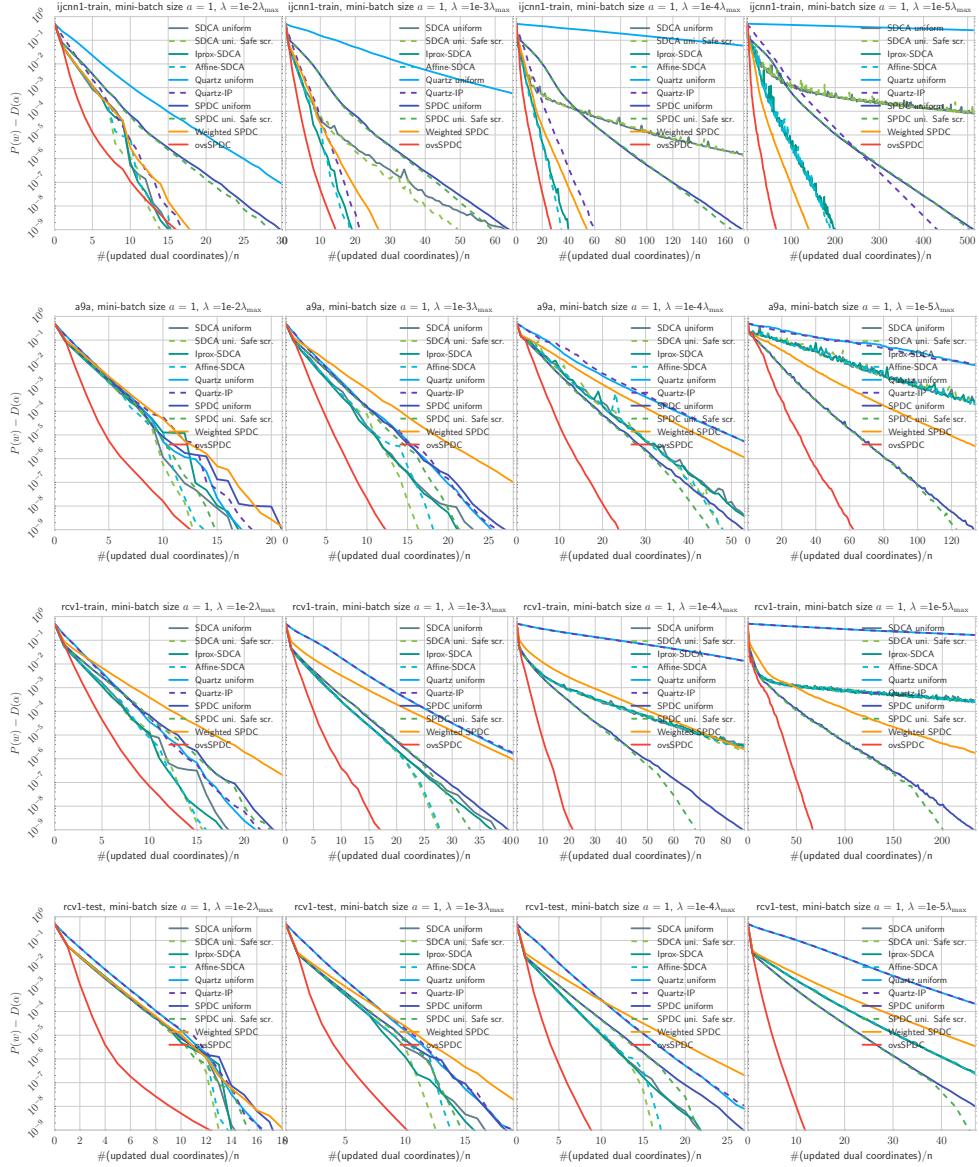

Figure 7: The illustration of the convergence speed of each method on `ijcnn1-train`, `a9a` (not normalized), `rcv1-train` and `rcv1-test` (normalized) in the setting is the same as §4.4.

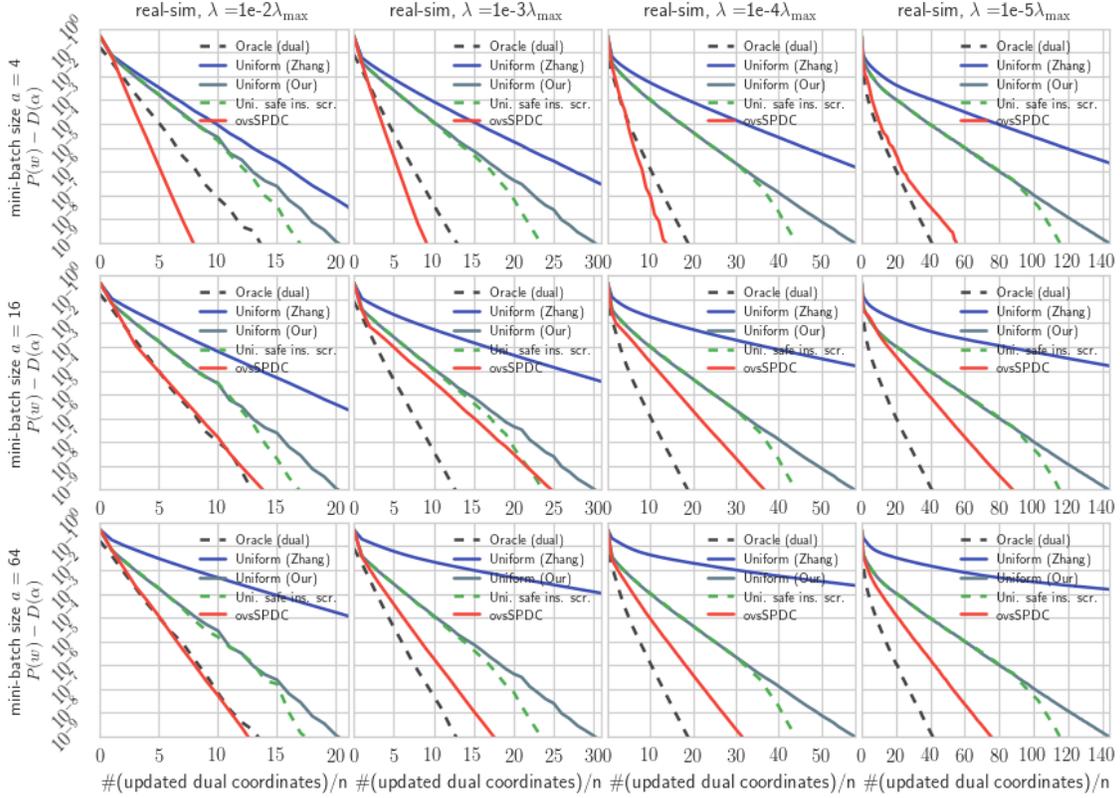

Figure 8: The illustration of the convergence speed of each method on `real-sim` in the setting is the same as §5.1 for $a = \{4, 16, 64\}$.

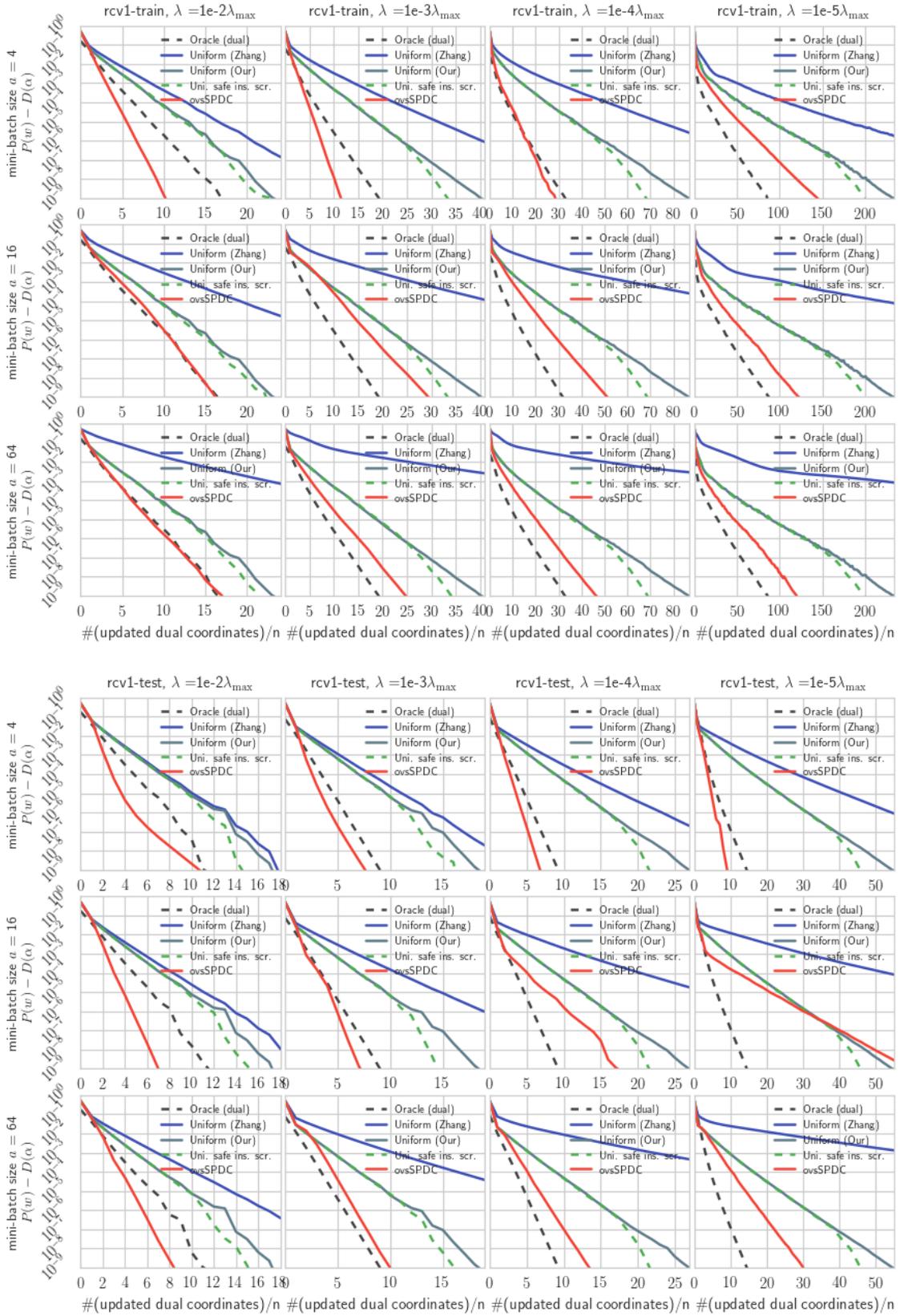

Figure 9: The illustration of the convergence speed of each method on `rcv1-train` and `rcv1-test` in the setting is the same as §5.1 for $a = \{4, 16, 64\}$.

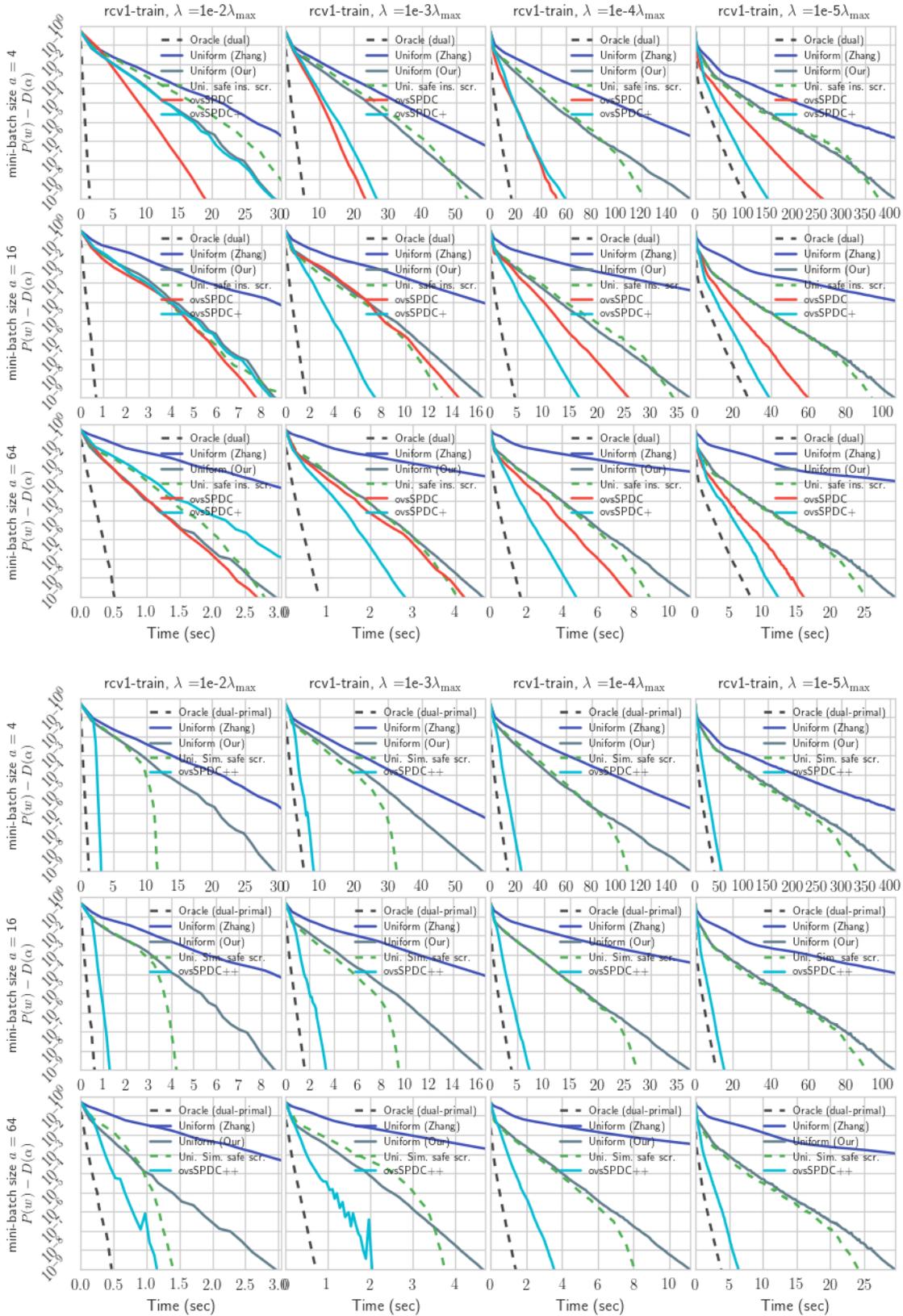

Figure 10: The illustration of the convergence speed of each method on rcv1-train in the setting is the same as §5.1 for $a = \{4, 16, 64\}$.

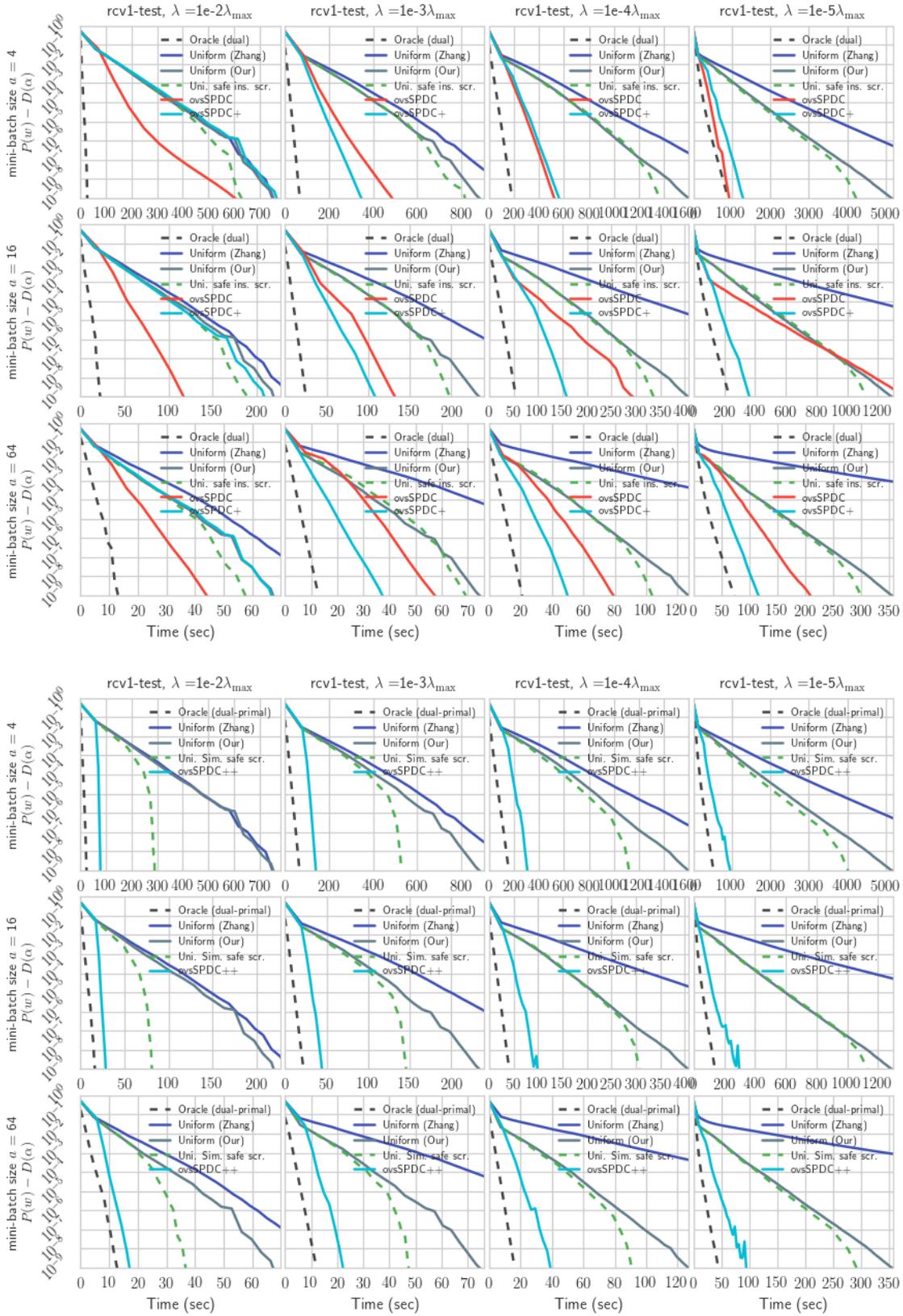

Figure 11: The illustration of the convergence speed of each method on `rcv1-test` in the setting is the same as §5.1 for $a = \{4, 16, 64\}$.

# G  Optimality Violations Sampling v.s. Heuristic Variants

In this appendix, we compare ovsSPDC proposed in §4.1 with two heuristic variants of SDCA through numerical experiments. AdaSDCA (Csiba et al., 2015) is theoretical methods, however AdaSDCA needs the calculation of $\kappa$ at each iteration and supports only the squared loss. In order to overcome these issues, the authors of (Csiba et al., 2015) proposed *AdaSDCA+*. AdaSDCA+ calculates $\kappa$ and sets $p$ with $p_i = \kappa_i \sqrt{\|x_i\|_2^2 + \lambda \gamma n}$ at every $n$ iterations. In the inner loop, AdaSDCA sets $p_i^{t+1} = p_i^t / m$, where $m$ is the hyper-parameter. *Empirical $\Delta$ SDCA*, which the authors of (Vainsencher et al., 2015) proposed, sets $p_i$ as depends on the $\alpha^{t+1} - \alpha_i^t$. We consider Quartz (Qu et al., 2015) with the sampling proposed in §4.1 which is called *ovsQuartz* because Quartz can prove the convergence with proper arbitrary sampling in the case of mini-batch size $a = 1$. The experimental setups are the same as those in §4.4. We consider the case of $a = 1$ because two heuristic variants does not support mini-batching. Figure 12 shows the results. AdaSDCA+ is faster than the others when $\lambda$ is large. ovsSPDC is stable and fast in all cases, especially when $\lambda$ is small.

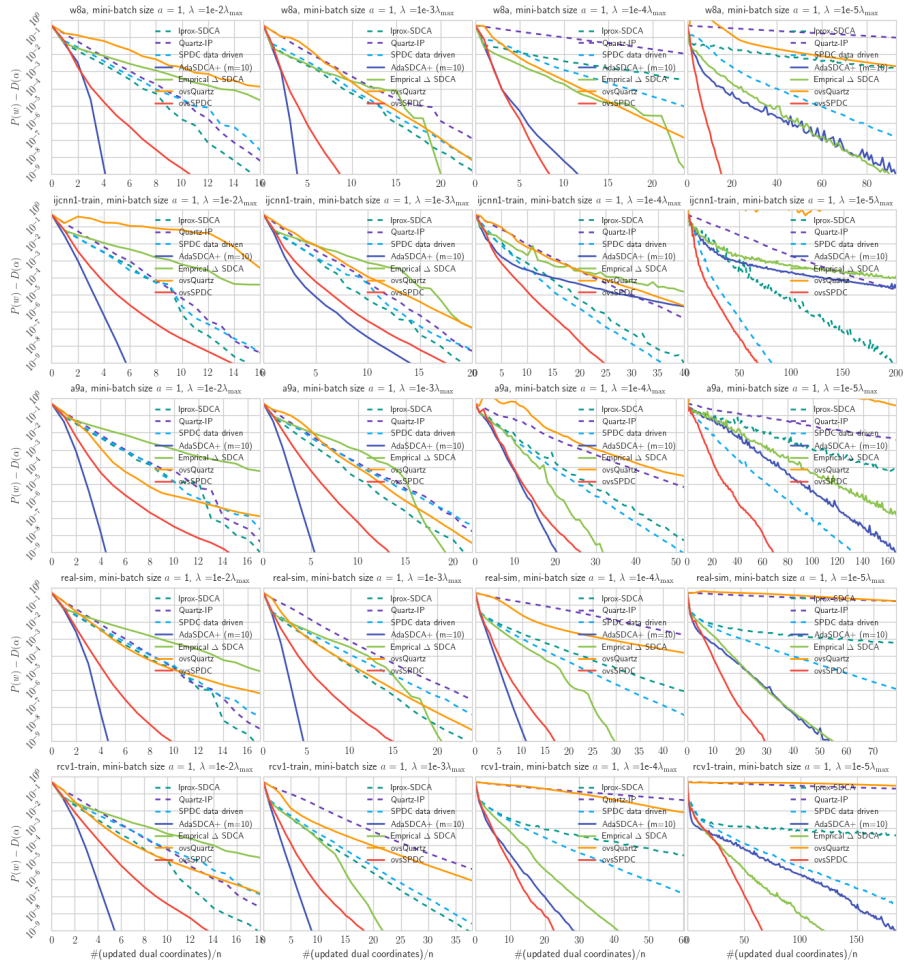

Figure 12: The illustration of the convergence speed of each method on each dataset.